\documentclass[sigconf]{acmart}

\AtBeginDocument{%
  \providecommand\BibTeX{{%
    \normalfont B\kern-0.5em{\scshape i\kern-0.25em b}\kern-0.8em\TeX}}}

\copyrightyear{2022} 
\acmYear{2022} 
\setcopyright{acmcopyright}
\acmConference[SIGIR '22]{Proceedings of the 45th International ACM SIGIR Conference on Research and Development in Information Retrieval}{July 11--15, 2022}{Madrid, Spain}
\acmBooktitle{Proceedings of the 45th International ACM SIGIR Conference on Research and Development in Information Retrieval (SIGIR '22), July 11--15, 2022, Madrid, Spain}
\acmPrice{15.00}
\acmDOI{10.1145/3477495.3531950}
\acmISBN{978-1-4503-8732-3/22/07}

\usepackage{subcaption}
\usepackage{multirow}
\usepackage{pifont}			
\usepackage{nicefrac}       
\usepackage{bm}
\usepackage{amsmath}
\usepackage{mathtools}
\usepackage[ruled,linesnumbered]{algorithm2e}
\usepackage{color}
\usepackage{colortbl}
\usepackage{hyperref}
\DeclareMathOperator*{\argmax}{arg\,max}
\DeclareMathOperator*{\argmin}{arg\,min}

\definecolor{Gray}{gray}{0.9}
\definecolor{Gray1}{gray}{0.8}

\newcommand{\ie}{\textit{i}.\textit{e}.}
\newcommand{\eg}{\textit{e}.\textit{g}.}

\acmSubmissionID{183}

\begin{document}

\title{CenterCLIP: Token Clustering for Efficient Text-Video Retrieval}

\author{Shuai Zhao}
\affiliation{%
\institution{CCAI, Zhejiang University}
\city{Hangzhou}
\country{China}
}
\email{zhaoshuaimcc@gmail.com}

\author{Linchao Zhu}
\affiliation{%
\institution{ReLER Lab, AAII, University of Technology Sydney}
\city{Sydney}
\country{Australia}
}
\email{zhulinchao7@gmail.com}

\author{Xiaohan Wang}
\affiliation{%
\institution{CCAI, Zhejiang University}
\city{Hangzhou}
\country{China}
}
\email{wxh1996111@gmail.com}

\author{Yi Yang}
\affiliation{%
\institution{\institution{CCAI, Zhejiang University}
\city{Hangzhou}
\country{China}
}
}
\email{yangyics@zju.edu.cn}


\begin{abstract}
   Recently, large-scale pre-training methods like 
    CLIP have made great progress in multi-modal research such as text-video retrieval.
    In CLIP, transformers are vital for modeling complex multi-modal relations.
    However, in the vision transformer of CLIP, the essential visual tokenization process, which produces discrete visual token sequences, generates many homogeneous
	tokens due to the redundancy nature of consecutive
	and similar frames in videos.
    This significantly increases computation
	costs and hinders the deployment of video retrieval models in web applications.
	In this paper, to reduce the number of redundant video tokens,
    we design a multi-segment token clustering algorithm
	to find the most representative tokens and drop the non-essential ones.
	As the frame redundancy occurs mostly in consecutive frames, we divide videos into multiple segments and conduct segment-level clustering.
	Center tokens from each segment are later concatenated into a new sequence, while their original spatial-temporal relations are well maintained.
	We instantiate two clustering algorithms to efficiently find deterministic medoids and iteratively partition groups in high dimensional space.
	Through this token clustering and center selection procedure,
	we successfully reduce computation costs by removing redundant visual tokens.
	This method further enhances segment-level semantic alignment between video and text representations, enforcing the spatio-temporal interactions of tokens from within-segment frames.
	Our method, coined as CenterCLIP, surpasses existing state-of-the-art
	by a large margin on typical text-video benchmarks, while reducing the training memory cost by 35\%
	and accelerating the inference speed by 14\% at the best case.
	The code is available at \href{{https://github.com/mzhaoshuai/CenterCLIP}}{{https://github.com/mzhaoshuai/CenterCLIP}}.
\end{abstract}

%


\maketitle

\begin{figure}[!t]
	\centering
	\includegraphics[width=0.46\textwidth]{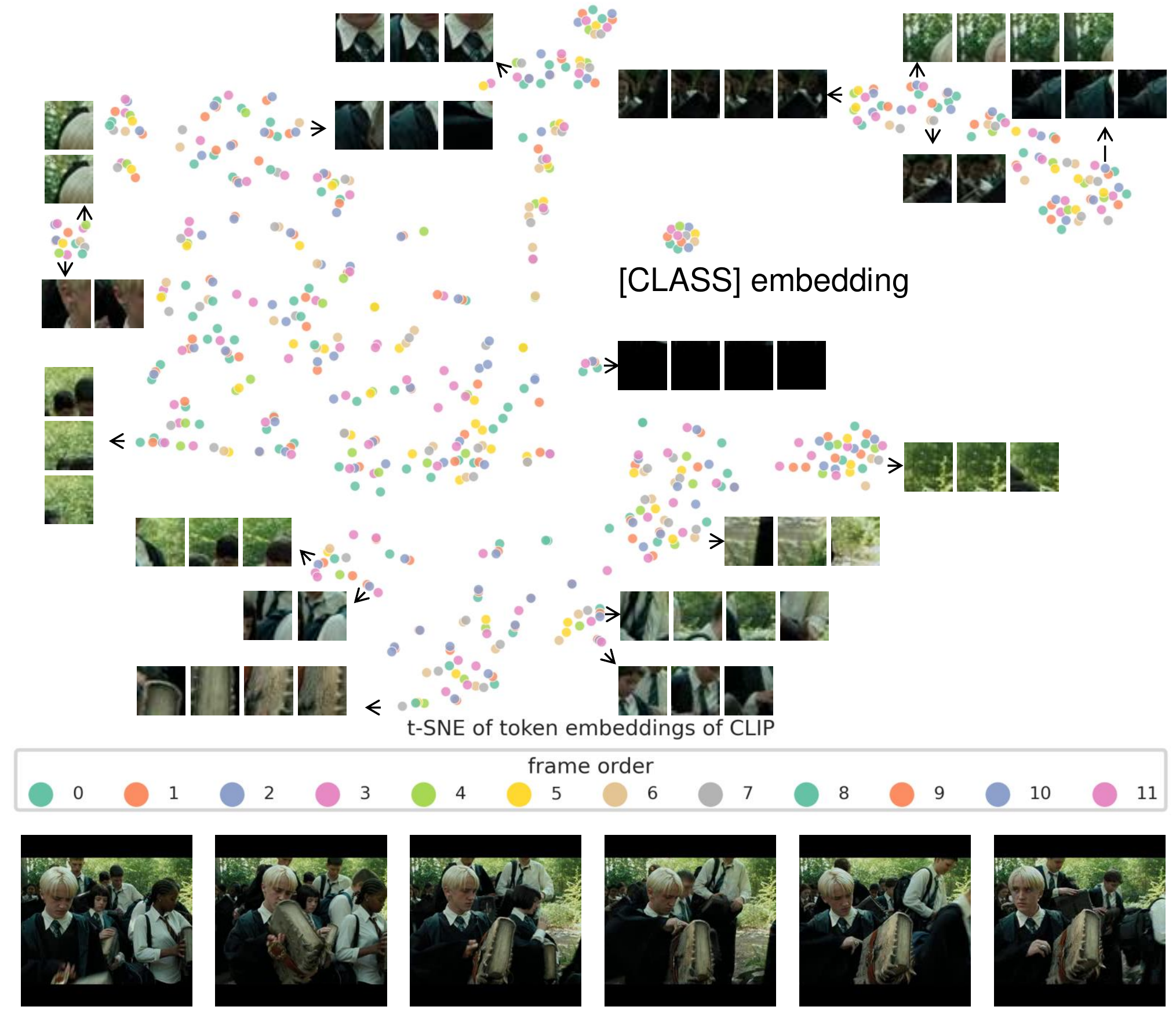}
	\caption{t-SNE~\cite{2008Visualizing} visualization of video token embeddings of CLIP.
		The shown similar image patches within a cluster are \textit{from different 
		temporal frames} in the same video.
		Best viewed in color with 300\% zoom.}
	\label{fig:tsne}
\end{figure}

\section{Introduction}

Text-video retrieval is less studied than the commonly known
text-image retrieval task as the intricate context of the video,
especially when the length of the video is very long or the
temporal variation of the video is large.
With the explosive growth of video content on mobile phones
and Internet during the past decade,
text-video retrieval becomes increasingly popular.
People also desire a better text-video retrieval system
as searching for videos of interest already becomes a part of
daily lives of most people.

Recently, with the success of large-scale 
contrastive language-image
pre-training methods like CLIP~\cite{2021-clip},
text-video retrieval also has made great progress.
To be specific, CLIP4clip~\cite{2021clip4clip} transfers the knowledge of CLIP to text-video retrieval tasks, surpassing the previous state-of-the-art methods by a large margin (\eg, more than 30\% improvement of the recall metric
on ActivityNet~\cite{caba2015activitynet}).
This demonstrates the power of billion-scale image-text pairs pre-training
via contrastive learning.
In CLIP, a vision transformer
\cite{DBLP:conf/nips/VaswaniSPUJGKP17,2021-vit} is adopted for visual representation 
learning.
Typically, in vision transformer, visual tokenization,
\ie, linear projection of non-overlapped image patches to an embedding space,
is a necessary component to produce discrete visual token sequences.
Then token sequences can be processed by the multi-head self-attention (MHSA)
in transformer blocks as the same manner of dealing with text sequences in the
original transformer~\cite{DBLP:conf/nips/VaswaniSPUJGKP17}.

When the input of the vision transformer becomes videos,
the visual tokenization procedure produces many homogeneous tokens
due to the redundancy nature in continuously changing frames.
In Figure~\ref{fig:tsne}, we extract the token embedding of CLIP from different frames in the same video and visualize them by t-SNE~\cite{2008Visualizing}.
From the visualization,
we can see those token embeddings from different frames form many tight clusters.
Image patches with similar texture features correspond to
immediate data points within a certain cluster.
It is also clear that
the number of clusters and the average number of tokens in clusters
are not small,
\ie, there are many similar token embedding in high-dimensional space.
As a result, repeated computation of these homogeneous tokens in CLIP inevitably
introduces a lot of unnecessary computation costs and hinders the training and deployment
of video retrieval models in web applications.
To resolve the above problem, in this work, we propose to
distinguish the most representative tokens,
\ie, the center token of each cluster in Figure~\ref{fig:tsne},
and only use these typical tokens for visual representation
learning as these tokens
contribute most to the discriminative
feature representation learning.

We introduce a multi-segment token clustering
algorithm to find the most representative tokens to reduce computation costs,
and achieve segment-level semantic alignment of video and text representation.
An input video is divided into multiple temporal segments. Each segment
contains the same number of consecutive frames.
Given the token embeddings of these frames,
a clustering algorithm is performed on each segment independently.
After clustering, only center tokens of clusters
are reserved and non-center tokens are dropped to 
avoid duplicated computation of similar tokens.
This significantly reduces computation costs.
Center tokens from the same temporal segment are
then concatenated into a new
visual sequence and arranged according
to their original spatial-temporal positions, \ie,
tokens whose image patches occur earlier in the video would appear at the earlier position of the new visual sequence.
Then the new visual sequence is processed by the standard transformer blocks.
This enables the model to learn segment-level video representation via attention among
tokens from within-segment frames.
These segment-level video representations are aligned with
the text through contrastive learning.

In this work, we introduce two instances of clustering algorithm
in multi-segment token clustering.
One is k-medoids equipped with a deterministic centroids initialization
method, \ie, KKZ initialization~\cite{1994_KKZ, DBLP:journals/ida/SuD07},
to ensure the clustering results are consistent through multiple runs.
A good initialization also helps the clustering algorithm converge fast.
The other is spectral clustering which suits high-dimensional 
data points clustering.
With our multi-segment token clustering
algorithm, CenterCLIP achieves state-of-the-art performance
on four common benchmarks:
MSR-VTT~\cite{xu2016msr},
MSVD~\cite{chen2011collecting},
LSMDC~\cite{rohrbach2015long},
and ActivityNet~\cite{caba2015activitynet}.
We achieve significant improvement of retrieval metrics on all these four datasets
compared to the baseline.
At the same time,
we achieve a decent reduction in memory cost
and speed up the inference process.
Specifically, on ActivityNet, we achieve a 35\% reduction in memory cost
and 14\% speedup of inference speed compared to the baseline.

\section{Related works}
\noindent\textbf{Contrastive Vision-Language Pre-Training.}
Since the success of derivative works
of Contrastive Language-Image
Pre-Training (CLIP)~\cite{2021-clip} in different areas
\cite{2021clip4clip, patashnik2021styleclip, bommasani2021opportunities,
shen2021much, xu2021videoclip},
visual representation learning under text supervision attracts
widespread attention.
Huge models pre-trained on billion-scale image-text pairs from web
like WenLan~\cite{huo2021wenlan},
Google's ALIGN~\cite{jia2021scaling}, and
Microsoft's Florence~\cite{yuan2021florence} emerged.
In the language-video understanding area,
there are similar works like
Frozen in Time~\cite{bain2021frozen} and HowTo100M~\cite{miech2019howto100m}.
However, the scale of language-video pre-training is much smaller than 
language-image pre-training as the former is much more expensive.
Following CLIP4clip~\cite{2021clip4clip},
we transfer the knowledge of CLIP to
the text-video retrieval task in this work.

\begin{figure*}[!t]
	\centering
	\includegraphics[width=0.97\textwidth]{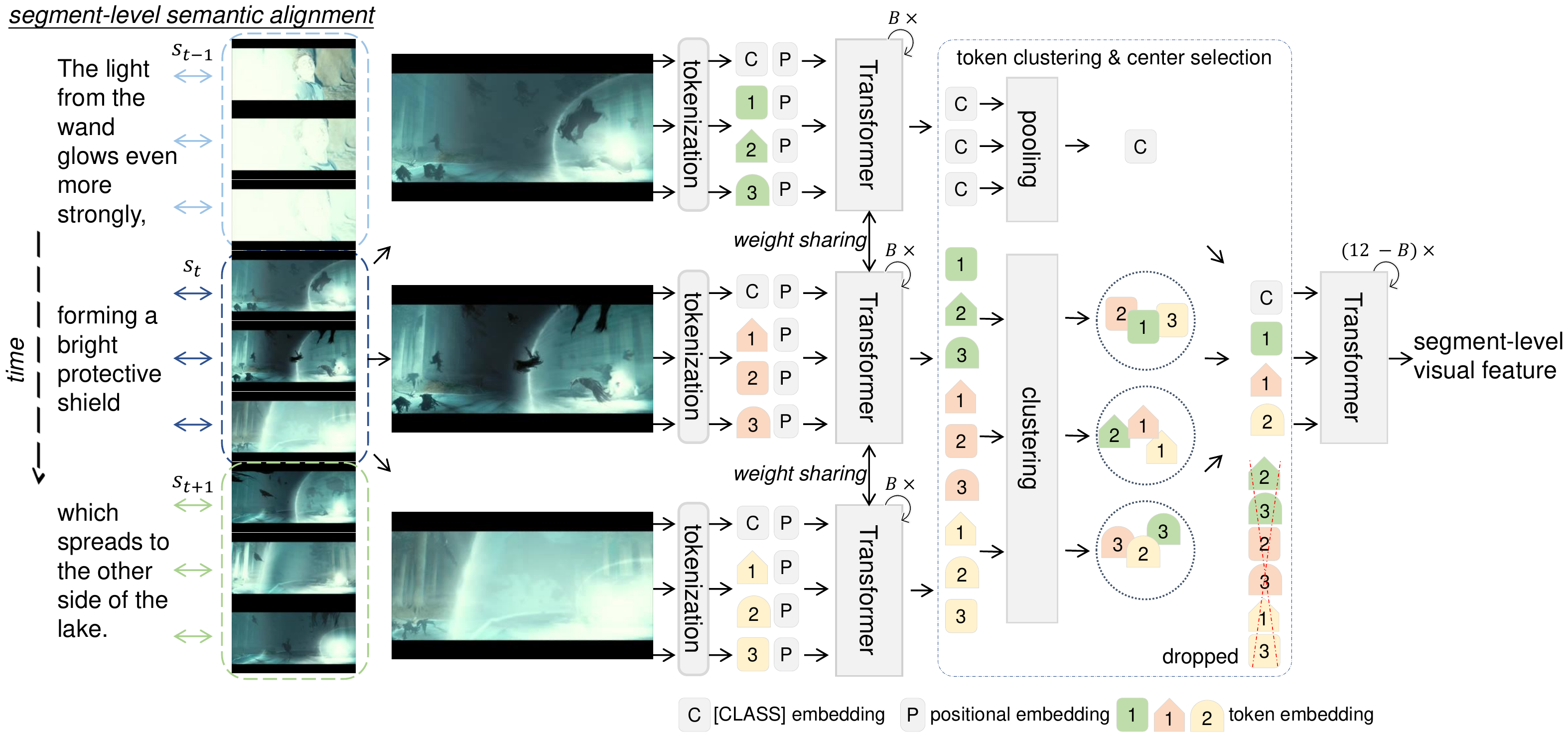}
	\caption{The overall framework of CenterCLIP and multi-segment clustering strategies.
    In this case, the video is divided into three segments and each contains three frames.
    Clustering is performed independently on tokens of
    each segment, and center tokens of all clusters
    from one segment are selected and
    concatenated into a new sequence.
    Via attention on this new sequence,
	the visual model is able to learn features that contain segment-level video semantics.
	This helps the whole text-video retrieval model to 
	achieve segment-level semantic alignment between video 
	and text while reducing computation costs.  }
	\label{fig:overall}
\end{figure*}

\noindent\textbf{Text-video Retrieval.}
Text-video retrieval is more complex than commonly
studied text-image retrieval as the additional temporal dimension
introduces complex context information.
Previously, standard language-video learning methods tend to
design dedicated fusion manners for cross-model
learning from offline extracted video and text features
\cite{yu2016video,yu2017end,le2020hierarchical,jang2017tgif, kaufman2017temporal,xu2017video}.
Recently, the paradigm of end-to-end large-scale pre-training plus task-specific finetune
becomes more and more popular for language-video understanding,
\eg, HowTo100M~\cite{miech2019howto100m}, MIL-NCE~\cite{miech2020end},
ActBERT~\cite{zhu2020actbert},
VideoBERT~\cite{sun2019videobert}, MMT~\cite{gabeur2020multi},
and HERO~\cite{li2020hero}.
These methods achieve promising results on many language-video tasks
and demonstrate the effectiveness of pre-training.
Our work is also in this line,
the difference is that we inherit the knowledge from CLIP~\cite{2021-clip},
which is pre-trained on image-text pairs rather than video-text pairs.

\noindent\textbf{Efficient Transformer.}
Recently, transformer becomes the unified model for many vision
and text tasks
\cite{DBLP:conf/nips/VaswaniSPUJGKP17,
radford2019language, 2021-vit, jaegle2021perceiver, liu2021swin}.
However, there are many time-consuming operations in transformers
such as self-attention and softmax operations.
Some works try to reduce the complexity of self-attention for very long sequences
or remove the softmax operation,
\eg, Performer~\cite{choromanski2020rethinking},
Linear Transformer~\cite{katharopoulos2020transformers},
Linformer~\cite{wang2020linformer}, Reformer~\cite{kitaev2020reformer},
Sparse Transformer~\cite{child2019generating},
Routing Transformer~\cite{roy2021efficient},
Longformer~\cite{beltagy2020longformer},
and Galerkin Transformer~\cite{Cao2021transformer}.
Very recently, in computer vision, people also notice
that not all tokens matter for the final performance
of the model.
To reduce the computation cost, researchers try to learn
a few most representative visual tokens
~\cite{ryoo2021tokenlearner,wu2020visual},
learn to rank all tokens and select the
most important ones~\cite{wang2021efficient},
and learn to mask the unimportant tokens
\cite{yin2021adavit, rao2021dynamicvit}.
Compared to these mentioned methods,
we are parameter-free
and introduce segment level semantic alignment of text
and video representation for text-video retrieval.

\section{Methods}

\subsection{Preliminary}
Given a video set $\mathcal{V}$ and text set $\mathcal{T}$,
the goal of text-video retrieval is to learn a score
function $\mathit{f}$, which gives a high similarity score
$\mathit{f}(v_i, t_i)$ if a video $v_i \in \mathcal{V}$ and
a text $t_i \in \mathcal{T}$ are highly relevant
and a low similarity score for an irrelevant video-text pair.
Then we can rank videos according to the query text
(text to video retrieval)
or rank texts according to the query video (video to text retrieval).

Following the typical multi-modal retrieval frameworks
\cite{2021-clip,lei2021less},
our text-video retrieval model is composed of a text encoder $\mathit{g}$
and a video encoder $\mathit{h}$.
Given a text $t_i$ and a video $v_i$,
$\mathit{g}(t_i)$ and $\mathit{h}(v_i)$
produce the \textit{normalized}
high-dimensional feature of the input,
where $\ell_2$ normalization is often considered in final feature encoding.
Then the similarity score of this text-video pair ($v_i$, $t_i$) is
\begin{align}
	\mathit{f}(v_i, t_i) = {\mathit{h}(v_i)^T \mathit{g}(t_i)}.
\end{align}
In training, a video-text pair ($v_i$, $t_i$) is treated
as the positive if $v_i$ and $t_i$ are corresponded.
All other instances of video or text in the mini-batch are treated as the negative.
The text and video encoders are optimized in an end-to-end manner via normalized
softmax loss~\cite{DBLP:conf/bmvc/ZhaiW19}.
The overall loss $\mathcal{L}$ is
the average of video-to-text classification loss ($\mathcal{L}_{v2t}$) and text-to-video classification loss ($\mathcal{L}_{t2v}$): 
\begin{align}
	\mathcal{L}_{v2t} &= - \frac{1}{N} \sum_{i}^{N} \log \frac{\exp(\mathit{h}(v_i)^T \mathit{g}(t_i) / \tau)}{
		\sum_{j=1}^{N} \exp(\mathit{h}(v_i)^T \mathit{g}(t_j) / \tau )},
\end{align}
\begin{align}
	\mathcal{L}_{t2v} &= - \frac{1}{N} \sum_{i}^{N} \log \frac{\exp(\mathit{g}(t_i)^T \mathit{h}(v_i) / \tau)}{
	\sum_{j=1}^{N} \exp(\mathit{g}(t_i)^T \mathit{h}(v_j) / \tau )}, \\
	\mathcal{L} &= \frac{1}{2}(\mathcal{L}_{v2t} + \mathcal{L}_{t2v}),
\end{align}
where $N$ is the mini-batch size and $\tau$ is the temperature to scale the logits.
It is worth noting that $\tau$ is crucial because
both $\mathit{h}(v_i)$ and $\mathit{g}(t_i)$ are  normalized.
We set it as a trainable parameter following the CLIP model.
During training, our model is initialized from the pre-trained weight of CLIP.
We describe the details of the text encoder and video encoder below.

\subsubsection{{\textbf{Text encoder}}}
We instantiate the text encoder using the text model of CLIP.
It is a transformer~\cite{DBLP:conf/nips/VaswaniSPUJGKP17}
with the architecture modifications described in BERT~\cite{radford2019language},
\ie, only encoder and no decoder.
A transformer model typically consists of repeated blocks~(layers)
of multi-head self-attention (MHSA) and feed-forward networks (FFN).
We use a transformer with 12 layers and 512 width with 8 attention heads,
where the width is the dimension of the query, key, and value feature.
The text tokenizer is a lower-cased byte pair encoding (BPE)~\cite{2016-bpe}
with a 49\,152 vocab size.
The text sequence is
padded with [\texttt{SOS}] and [\texttt{EOS}] tokens.
[\texttt{SOS}] and [\texttt{EOS}]
is padded at the beginning and end of the text sequence, respectively.
The final text feature representation is the activation from the last layer of the transformer that corresponds to the [\texttt{EOS}] token.
This text representation is later normalized by layer normalization and linearly projected into the joint video-text embedding space.

\subsubsection{\textbf{Video encoder}}
Our video encoder is a vision transformer (ViT),
which first successfully applied transformers in vision tasks.
The architecture of ViT is the same as the transformer in 
natural language processing,
except ViT introduces an additional visual tokenization process to convert
images into discrete sequences.
When feeding images or videos into a ViT, we first
convert the non-overlapped image patches into visual sequences,
where a [\texttt{CLASS}] token is prepended to the beginning of sequences
as BERT~\cite{radford2019language}.
Then the output of [\texttt{CLASS}] token at the final layer is extracted
as the visual representation.
In this work, we adopt a 2D linear projection
to project image patches of different frames into an
embedding space independently
following the practice of CLIP4clip~\cite{2021clip4clip}.
For convenience, we name this linear transformation process as
\textit{visual tokenization}.
Generally, we use a ViT-B/32 model~\cite{2021-vit} with 12 layers and
512 width with 8 attention heads.
ViT-B/32 means the non-overlapped input image patch size is $32 \times 32$.

When applying the visual tokenization process to videos,
it inevitably produces many redundant tokens as shown in Figure~\ref{fig:tsne}.
Generally, an input video $v_i$
consists of many temporal related frames:
$v_i = \{v_i^1, v_i^2, \ldots,  v_i^{|v_i|}\}$,
where $|v_i|$ is the number of frames in $v_i$.
After visual tokenization,
if each frame produces $L$ tokens, 
the number of visual tokens is $L|v_i|$
(do not consider [\texttt{CLASS}] token).
It shows that the number of visual tokens is linear to
the number of tokens per frame ($L$) and the video length.
Given an input frame with a size of $224 \times 224$,
$L=49$ for the ViT-B/32 model and $L=196$ for the ViT-B/16 model. 
With a larger $L$, the number of visual tokens becomes much larger.
When performing text-video retrieval on long videos,
the total number of tokens for a video is large.
For example, videos in the ActivityNet~\cite{caba2015activitynet}
dataset usually have a few minutes duration. In this case,
$L|v_i|$ will be easily larger than 1\,000.

The redundant tokens considerably increase computation costs.
To make the training and inference of text-video retrieval models more efficient,
we propose to use clustering algorithms to
find the most representative token embeddings.
This process significantly reduces the number of tokens while maintaining
the most valuable information of original tokens.
After clustering, we only reserve the center tokens and
remove other non-center tokens.
The reserved tokens contain most of the information about the video
and it is sufficient for text-video retrieval.
We describe our multi-segment token clustering in the next section.

\subsection{Multi-segment Token Clustering}
The overall framework of our video encoder can be found in Figure~\ref{fig:overall}.
We perform a multi-segment clustering strategy
on visual tokens from a certain temporal segment.
This is based on the assumption that neighbor frames are more
likely to be the same;
then tokens of these similar neighbor frames are more
possible to be redundant.
Our multi-segment token clustering method empowers the model
to achieve segment-level semantic alignment of text and video representations.
Previously, CLIP and CLIP4clip adopt the average of frame features or
the late fusion of frame features as the video representation.
However, the former loses the temporal information,
and the latter is a post-processing step and loses the detail temporal
variations at the early stage of the transformer.
By clustering across multiple frames within a segment
at an early or middle stage,
image patches from different temporal positions can interact with each other
via the self-attention mechanism.

Specifically, a video sequence
$\{v_i^1, v_i^2, \ldots,  v_i^{|v_i|}\}$ is divided into $S$ segments
$\{s_i^1, s_i^2, \ldots, s_i^S\}$.
Each segment contains $ \frac{|v_i| }{S}$ frames and $\frac{L|v_i| }{S}$ tokens.
Then we perform token clustering on these $\frac{L|v_i| }{S}$ tokens segment-wise,
namely, clustering for each segment independently.
Then the centers of all clusters
from one segment,
\ie, center tokens, are selected and
other non-center tokens are simply dropped.
These center tokens are concatenated and arranged according to
their original relative spatial-temporal position.
Center tokens from the upper-left position and early frames are
at the beginning of the new token sequence.
Center tokens from the bottom-right position and late frames are
at the rear-end of the new token sequence.

Multi-segment token clustering algorithm makes our vision model achieve
segment-level temporal modeling and be able to capture the
detailed temporal variation of video frames.
This allows our methods to achieve segment-level alignment of the text $t_i$
and the video $v_i$ consisted of segments $\{s_i^1, s_i^2, \ldots, s_i^S\}$:
\begin{align}
	f(v_i, t_i) = \frac{1}{S}\sum_{j=1}^{S}h(s_i^j)^T g(t_i).
\end{align}

The multi-segment token clustering method has at least two advantages:
\textit{(1)} reducing computation costs by cutting down the number of tokens;
\textit{(2)} achieving segment-level semantic alignment
of text and video representations via attention among
tokens from different frames within the same temporal segment.
As shown in Figure~\ref{fig:overall}, assuming we perform token clustering
right after the $B$-th transformer block and the number of clusters is $K$
(ignore [\texttt{CLASS}] token after pooling),
this means the length of the input sequence length of the following $(12-B)$
transformer blocks become $K$.
Generally, $\frac{L|v_i| }{S} >> K$, obviously,
computational costs are largely reduced.
It is worth noting that the clustering module can be
inserted at any place of ViT and
the clustering procedure can be performed for any times.
Clustering at an early stage reduces more computation costs.
Next, we discuss two clustering methods used in the
multi-segment token clustering algorithm.

\subsubsection{\textbf{k-medoids++}}
In this section,
we introduce the first instance of the token clustering method: k-medoids++,
a variety of commonly known k-means algorithm.
Given a set of tokens $\{x_{1}, ... , x_{m} \} \in \mathbb{R}^d$,
where $d$ is the transformer width and $m = \frac{L|v_i| }{S}$
is the number of tokens in a temporal segment,
the goal of k-means is to partition the $m$ observations into $K~(\le m)$ sets,
\ie, $\bm{\mathcal{C}} = \{\mathcal{C}_1, \mathcal{C}_2, \ldots,  \mathcal{C}_K \}$,
so as to minimize the within-cluster sum of squares:
$\mathop{\arg\min}_{\bm{\mathcal{C}}} \sum_{i=1}^{K}\sum_{x \in \mathcal{C}_i} \lVert x - \mu_i\rVert_2^2$,
where $\mu_i$ is the mean of points in $\mathcal{C}_i$, \ie, centroid.
Normal k-means contains 4 steps:
\begin{itemize}
\item[1.] Initialize cluster centroids $\mu_1$, $\mu_2$, $\ldots$, $\mu_K$ $\in \mathbb{R}^d$ randomly;
\item[2.] For every $i$, set $p_i \coloneqq \argmin_{j} \lVert x_i - \mu_j\rVert_2^2$;
\item[3.] For every $j$, set $\mu_j  \coloneqq \frac{\sum_{i=1}^{m}\mathbf{1}\{p_i = j\} x_i }{\sum_{i=1}^{m} \mathbf{1}\{p_i = j\}}$; $\mathbf{1}\{\cdot\}$ equals to $1$ if and only if the inner condition is true;
\item[4.] Repeat step 2 and 3 until convergence.
\end{itemize}

One disadvantage of the normal
k-means is that clustering results
are sensitive to the centroids initialization.
Bad initialization may lead to the collapse of the clustering.
Random initialization is also not suitable for retrieval,
as we would obtain inconsistent retrieval
results when querying multiple times. 
Therefore, we need a deterministic centroids initialization method.
In this work, we adopt the KKZ initialization
method~\cite{1994_KKZ, DBLP:journals/ida/SuD07}.
The algorithm is shown in Algorithm~\ref{algo-kkz}.
It first chooses the point with the maximum $\ell_2$-norm as the first centroid,
then chooses the point with the maximum distance to the existing centroids
as the next centroids.
The algorithm is simple but effective~\cite{1994_KKZ, DBLP:journals/ida/SuD07}.
It makes k-means deterministic and accelerates its convergence speed.

In our token clustering process, we use medoids rather than centroids.
Namely, we choose the nearest point to centroids as the seed point in
steps 2\&3 in the normal k-means methods.
This is because the semantic of mean pooling
representation of tokens within a cluster may shift away
from the exact token embeddings.
Combining k-medoids and KKZ initialization,
we get k-medoids++ --- the name philosophy follows k-mean++~\cite{DBLP:conf/soda/ArthurV07}.
The complexity of k-medoids++ is $\mathcal{O}(mKdI)$,
where $I$ is the iteration upper bound
and $\mathcal{O}(d)$ for computing the distance between two points in $\mathbb{R}^d$.

\begin{algorithm}[t]
	\caption{KKZ initialization for k-means~\cite{1994_KKZ}}
	\label{algo-kkz}
	\KwIn{tokens $\{x_{1}, ... , x_{m} \} \in \mathbb{R}^d$,
		cluster number $K$}
	\KwOut{centroids $\{\mu_1$, $\mu_2$, $\ldots$, $\mu_K\}$ $\in \mathbb{R}^d$}
 	$\mu_1 \leftarrow \argmax_{x_i} \lVert x_i \rVert_2$\;
 	
 	\For{$i\leftarrow 2$ \KwTo $K$}
	{
		\For{$j\leftarrow 1$ \KwTo $m$}
		{
			\For{$k\leftarrow 1$ \KwTo $i$}
			{	
			\tcp{calculate distance between $x_j$ and $\mu_k$}
			$d_{j,k} \leftarrow \mathtt{distance}(x_j, \mu_k)$\;
			}
			$d_j \leftarrow \min_k d_{j, k}$\;
		}
		$\mu_i \leftarrow \argmax_{x_j} d_j$\;
	}
\end{algorithm}

\subsubsection{\textbf{Spectral clustering}}
K-means is suitable for spherical data clusters.
However, our data points are
in a high dimension space $\mathbb{R}^d$ ($d=512$ in most cases), and
the data distribution is unknown.
The shape of data clusters may not be spherical.
To resolve this underlying problem,
we further introduce spectral clustering into the token clustering process.
Spectral clustering is a graph partition method that aims to maximize the
weights of connections within groups and minimize the weights of connections
between groups.
It first needs to construct the graph $G=(X, E)$,
where $X$ is the vertex set and each vertex
represents a data point $x$, $E$ is the edge set and
each edge denotes the (weighted) connection between two vertices.
In this work, we use the normalized spectral clustering algorithm described in \cite{Ng01onspectral}.
The algorithm contains 5 steps:
\begin{itemize}
	\item[1.] Construct similarity graph. Let $W$ be its weighted adjacency matrix, $D$ be the degree matrix;
	\item[2.] Compute normalized Laplacian $L_{sym} = D^{-\frac{1}{2}}(D - W)D^{-\frac{1}{2}}$;
	\item[3.] Compute the first $K$ eigenvectors $\mu_1,\ldots,\mu_k$ of $L_{sym}$
	which correspond to the first $K$ least eigenvalues;
	\item[4.] Let $U=[\mu_1,\ldots,\mu_k]$; Normalize each row of $U$ to have norm of $1$, generally, $\ell_2$ norm is used;
	\item[5.] Consider each row of $U$ as a new data point, apply k-means to these data points.
\end{itemize}
\begin{algorithm}[t]
	\caption{sign flip for SVD~\cite{signflip1}}
	\label{algo-sign}
	\KwIn{$L \in \mathbb{R}^{m\times m}$,
		truncated singular value decomposition $(U, \Sigma, V)$
		of $L$, $U = [u_1, \ldots, u_K] \in \mathbb{R}^{m\times K}$}
	\KwOut{$U^{\prime}= [u_1^\prime, \ldots, u_K^\prime]$ with appropriate signs}
	\For{$k\leftarrow 1$ \KwTo $K$}
	{
		$Y = L - \sum_{i=1, i \ne k}^{K} \sigma_i u_i v_i^T $\;
		\tcc{$\mathtt{sign}(\cdot)$ return the sign of input,
			$Y_{\cdot, j}$ denote the $j$-th column of $Y$}
		$s_k = \sum_{j=1}^{m}\mathtt{sign}(u_k^TY_{\cdot, j})(u_k^TY_{\cdot, j})^2$\;
	}
	\For{$k\leftarrow 1$ \KwTo $K$}
	{
		$u_k^\prime = \mathtt{sign}(s_k)u_k$\;
	}
\end{algorithm}
The above algorithm first performs dimension reduction and then data clustering.
In this work, we use SVD to solve the eigenvectors of $L_{sym}$ as $L_{sym} = L_{sym}^T$.
A sign correct algorithm is further introduced to resolve the sign
ambiguity in SVD as the direction of points after dimension
reduction also matters for some distance metrics, \eg, $\ell_2$ distance.
The main idea of this sign correct algorithm is to make the 
direction of singular vectors aligns with
the majority direction of data points~\cite{signflip1},
\ie, the sign of the sum of the inner product of singular vectors
and data points should be positive.
Here we describe the special case of this algorithm for
symmetric matrix in Algorithm~\ref{algo-sign}.
The complexity of spectral clustering is $\mathcal{O}(mK^2I + m^3)$, where $\mathcal{O}(m^3)$
for step 1-4 and $\mathcal{O}(mK^2I)$ for step 5.
For more details about spectral clustering refer to the tutorial~\cite{von2007tutorial}.


\begin{table*}[!t]
	\centering
	\scalebox{0.82}{
		\begin{tabular}{lcc|ccccc|ccccc}
			\toprule 
			\multirow{2}{*}{\centering Method}
			& MeM.  & Speed
			& \multicolumn{5}{c}{Text $\rightarrow$ Video}
			& \multicolumn{5}{c}{Video $\rightarrow$ Text } \\
			
			~& GB & ms
			& R@1$\uparrow$   & R@5$\uparrow$ & R@10$\uparrow$ & MdR$\downarrow$ & MnR$\downarrow$ 
			& R@1$\uparrow$   & R@5$\uparrow$ & R@10$\uparrow$ & MdR$\downarrow$ & MnR$\downarrow$  \\
			\midrule 

			CE~\cite{liu2019use}	
			& - & - 
			& 19.8 & 49.0 & 63.8 & 6 & - 
			&  - & -   &-   &-   & -\\ 		
			
			TT-CE+~\cite{croitoru2021teachtext}
			& - & - 
			& 25.4 & 56.9 & 71.3 & 4 & - 
			& 27.1 & 55.3 & 67.1 & 4 & - \\			
			
			Frozen in Time~\cite{bain2021frozen}
			& - & - 
			& 33.7 & 64.7 & 76.3 & 3 & - 
			& - & - & - & - & - \\			
			
			CLIP zero-shot
			& - & - 
			& 37.0 & 64.1 & 73.8 & 3 & -
			& 59.9 & 85.2 & 90.7 & 1 & - \\
			
			CLIP4clip~(meanP)~\cite{2021clip4clip}
			& 20.8 & 24.4
			& 46.2 & 76.1 & 84.6 & 2 & 10.0 
			& 56.6 & 79.7 & 84.3 & 1 & 7.6 \\

			CLIP4clip~(seqTransf)
			& - & - 
			& 45.2 & 75.5 & 84.3 & 2 & 10.3
			& 62.0 & 87.3 & 92.6 & 1 & 4.3 \\
			\midrule 
			
			baseline (CLIP4clip~(meanP), ViT-B/32)
			& 20.8 & 24.4 
			& 45.9 & 74.9 & 84.7 & 2 & 10.4
			& 51.0 & 76.3 & 82.2 & {1} & 9.1 \\
			
			CenterCLIP~(k-medoids++, $B_6-4, 49$)
			& 15.0 & 22.9 
			& 47.6 & 77.5 & 86.0 & 2 & 9.8 
			& 54.2 & 78.4 & 84.9 & 1 & 7.6 \\
			
			CenterCLIP~(k-medoids++, $B_6-3, 49$)
			& \cellcolor{Gray}{14.2} & \cellcolor{Gray}{22.9}
			& 47.3 & 76.8 & 85.6 & 2 & 9.9
			& 57.9 & 83.6 & 90.5 & 1 & 5.2  \\
			
			CenterCLIP~(spectral, $B_6-4, 49$)
			& 14.9 & 40.8 
			& 47.4 & 76.5 & 85.2 & 2 & 9.7 
			& 62.7 & 88.1 & 92.8 & 1 & 4.1 \\
			
			CenterCLIP~(spectral, $B_6-3, 49$)
			& 14.2 & 43.6 
			& 47.3 & 76.9 & 86.0 & 2 & 9.7
			& 63.5 & 86.4 & 92.6 & 1 & 3.8  \\
			
			\midrule
			baseline (CLIP4clip~(meanP), ViT-B/16)
			& 25.7 & 59.6
			& 49.6 & 79.5 & 88.0 & 2 & 8.6
			& 62.7 & 83.9 & 89.4 & {1} & 6.1 \\			

			CenterCLIP~(k-medoids++, $B_6-4, 160$)
			& 17.6 & 86.5 
			& \cellcolor{Gray}50.6 & \cellcolor{Gray}80.3 & \cellcolor{Gray}88.4 & \cellcolor{Gray}1 &  \cellcolor{Gray}8.4 
			& \cellcolor{Gray}68.4 & \cellcolor{Gray}90.1 & \cellcolor{Gray}95.0 & \cellcolor{Gray}1 & \cellcolor{Gray}3.0 \\						
			\bottomrule
		\end{tabular}
	}
	\caption{ Results on MSVD.  
		MeM. is the average GPU memory cost when training on 2 and 8 Tesla V100 GPUs for ViT-B/32 and ViT-B/16, respectively.
		Speed is the inference time per video during evaluation on a Tesla V100 GPU. 
	}
	\label{tab:result_MSVD}
\end{table*}

\begin{table*}[t]
	\centering
	\scalebox{0.83}{
		\begin{tabular}{lcc|ccccc|ccccc}
			\toprule 
			\multirow{2}{*}{\centering Method}
			& MeM.  & Speed
			& \multicolumn{5}{c}{Text $\rightarrow$ Video}
			& \multicolumn{5}{c}{Video $\rightarrow$ Text } \\
			
			~& GB & ms
			& R@1$\uparrow$   & R@5$\uparrow$ & R@10$\uparrow$ & MdR$\downarrow$ & MnR$\downarrow$ 
			& R@1$\uparrow$   & R@5$\uparrow$ & R@10$\uparrow$ & MdR$\downarrow$ & MnR$\downarrow$  \\
			\midrule 

			FSE~\cite{zhang2018cross}	
			& - & - 
			& 18.2 & 44.8 & - & 7.0 & - 
			& 16.7 & 43.1 & - & 7.0  & -\\ 	
			
			CE~\cite{liu2019use}	
			& - & - 
			& 18.2 & 47.7 & - & 6.0 & 12.1 
			& 17.7 & 46.6  & -  & 6.0  & 24.4 \\ 	
			
			CMGSD~\cite{he2021improving}	
			& - & - 
			& 24.2 & 56.3 & - & 4.0 & - 
			& 24.6 & 56.8 & - & 4.0 & - \\
			
			MMT~\cite{gabeur2020multi}	
			& - & - 
			& 28.7 & 61.4 & - & 3.3 & 16.0 
			& 28.9 & 61.1 & - & 4.0 & 17.1 \\	
			
			TT-CE+~\cite{croitoru2021teachtext}
			& - & - 
			& 23.5 & 57.2 & - & 4.0 & - 
			& 23.0 & 56.1 & - & 4.0 & - 	\\		
			
			T2VLAD~\cite{wang2021t2vlad}
			& - & - 
			& 23.7 & 55.5 & - & 4.0 & - 
			& 24.1 & 56.6 & - & 4.0 & - \\		

			SSB~\cite{patrick2020support}
			& - & - 
			& 29.2 & 61.6 & - & 3.0 & - 
			& 28.7 & 60.8 & - & 2.0 & - 	\\	
			
			CLIP zero-shot
			& - & - 
			& 21.7 & 46.0 & 59.6 & 7.0 & 39.7
			& 17.9 & 40.8 & 54.2 & 8.0 & 43.3  \\

			CLIP4clip~(meanP)~\cite{2021clip4clip}
			& 25.0 & 82.0 
			& 40.5 & 72.4 & - & 2.0 & 7.4 
			& 42.5 & 74.1 & 85.8 & 2.0 & 6.6 \\
			
			CLIP4clip~(seqTransf)
			& - & - 
			& 40.5 & 72.4 & - & 2.0 & 7.5
			& 41.4 & 73.7 & 85.3 & 2.0 & 6.7 \\
			\midrule 
			
			baseline (CLIP4clip~(meanP), ViT-B/32)
			& 25.0 & 82.0 
			& 41.8 & 73.9 & 84.7 & 2.0 & 7.3
			& 42.8 & 73.8 & 85.3 & 2.0 & 6.9 \\

			CenterCLIP~(k-medoids++, $B_6-{15}, 49$)
			 & 16.8 & 71.3
			& 43.9 &  75.3 & 85.2 & 2.0 & 7.0
			& 44.2 & 75.0 & 86.1 & 2.0 & 6.8  \\
			
			CenterCLIP~(k-medoids++, $B_6-{12}, 49$)
			& \cellcolor{Gray}16.2 & \cellcolor{Gray}70.4 
			& 43.5 & 75.0 & 85.9 & 2.0 & 6.9 
			& 44.5 & 75.3 & 86.0 & 2.0 & 6.7 	  \\
			
			CenterCLIP~(spectral, $B_6-{20}, 49$)
			& 17.7 & 162 
			& 43.5 & 75.1 & 85.4 & 2.0 & 6.9 
			& 44.1 & 75.1 & 86.0 & 2.0 & 6.7  \\
			
			CenterCLIP~(spectral, $B_6-{15}, 49$)
			& 16.8 & 174 
			& 43.9 & 74.6 & 85.8 & 2.0 &  6.7
			& 44.5 & 75.7 & 86.2 & 2.0 & 6.5  \\
			\midrule

			CenterCLIP~(k-medoids++, $B_6-{15}, 160$, ViT-B/16)
			& 23.0 & 419 
			& \cellcolor{Gray}46.2 & \cellcolor{Gray}77.0
			& \cellcolor{Gray}87.6 & \cellcolor{Gray}2.0
			& \cellcolor{Gray}5.7 
			& \cellcolor{Gray}46.7 & \cellcolor{Gray}77.1 
			& \cellcolor{Gray}88.0 & \cellcolor{Gray}2.0 
			& \cellcolor{Gray}5.5  \\	

			\bottomrule
		\end{tabular}
	}
	\caption{ Results on ActivityNet. 
		MeM. is the average GPU memory cost when training on 8 and 
		32 Tesla V100 GPUs. Baseline with ViT-B/16 OOM on 32 Tesla V100 GPUs.
		Speed is the inference time per video during evaluation on a Tesla V100 GPU. 
	}
	\label{tab:result_ACT}
\end{table*}

\section{Experiments}
\subsection{Experimental Details}
\noindent
\textbf{Datatest.}
We validate our model on four datasets: MSR-VTT~\cite{xu2016msr}, MSVD~\cite{chen2011collecting}, LSMDC~\cite{rohrbach2015long}, and ActivityNet~\cite{ caba2015activitynet}.
To save computational costs, the shorter side of videos
are resized to 224 and the frame per second (fps) is set to 3.
~\textbf{(a)}
\textbf{MSR-VTT} contains 10\,000 videos with a length
ranges from 10~\textasciitilde~32 seconds and 200\,000 captions.
We use two types of data splits, \textsf{training-7K} and \textsf{training-9K},
to compare with baselines.
The \textsf{training-7K} follows the data splits from HowTo100M~\cite{miech2019howto100m}
and the \textsf{training-9K} follows the data splits from \cite{gabeur2020multi}.
The test data in both splits is `\textsf{test 1k-A}',
which contains 1\,000 video-text pairs following JSFusion \cite{yu2018joint}.
If we do not specify, we use \textsf{training-9K} as the default.
~\textbf{(b)}
\textbf{MSVD} contains 1\,970 videos
with a duration ranges from 1~\textasciitilde~62 seconds.
Train, validation, and test splits contain 1\,200, 100, and 670 videos, respectively.
Each video has approximately 40 associated sentences in English.
~\textbf{(c)}
\textbf{LSMDC} is comprised of 118\,081
videos that ranges from 2\textasciitilde30 seconds.
The videos were extracted from 202 movies.
The validation set contains 7\,408 videos.
The 1\,000 videos in the test set are from movies independent
from the training and validation splits.
~\textbf{(d)}
\textbf{ActivityNet}~\cite{heilbron2014collecting, caba2015activitynet}
consists of 20\,000 YouTube videos, and some of them are minutes long.
We follow~\cite{zhang2018cross,gabeur2020multi} to concatenate
all the descriptions of a video to form a paragraph and evaluate the
model with video-paragraph retrieval on the \textsf{val1} split.

\begin{table*}[tbp]
	\centering

	\begin{subtable}{1.0\linewidth}
	\centering
	\scalebox{0.83}{
		\begin{tabular}{lcc|ccccc|ccccc}
			\toprule 
			\multirow{2}{*}{\centering Method}
			& MeM.  & Speed
			& \multicolumn{5}{c}{Text $\rightarrow$ Video}
			& \multicolumn{5}{c}{Video $\rightarrow$ Text } \\
			
			~& GB & ms
			& R@1$\uparrow$   & R@5$\uparrow$ & R@10$\uparrow$ & MdR$\downarrow$ & MnR$\downarrow$ 
			& R@1$\uparrow$   & R@5$\uparrow$ & R@10$\uparrow$ & MdR$\downarrow$ & MnR$\downarrow$  \\
			\midrule 
			
			CLIP4clip~(meanP)~\cite{2021clip4clip}
			& 20.8 & 24.4
			& 42.1 & 71.9 & 81.4 & {2} & 15.7 
			& - & - & - & - & - \\
			
			CLIP4clip~(seqTransf)
			& - & - 
			& 42.0 & 68.6 & 78.7 & 2 & 16.2
			& - & - & - & - & - \\
			\midrule 
			
			baseline (CLIP4clip~(meanP), ViT-B/32)
			& 20.8 & 24.4 
			& 42.4 & 69.2 & 79.4 & {2} & 17.3
			& 42.2 & 68.8 & 78.8 & {2} & 12.5 \\
			
			CenterCLIP~(k-medoids++, $B_6-4, 49$)
			& 15.0 & 22.9 
			& 43.7 & {71.3} & 80.8 & 2 & 16.9 
			& 41.8 & 68.9 & 77.9 & {2} & {13.3} \\
			
			CenterCLIP~(k-medoids++, $B_6-3, 49$)
			& \cellcolor{Gray}{14.2} & \cellcolor{Gray}{22.9}
			& 43.5 & 68.5 & 79.7 & {2} & 17.7
			& 40.9 & 68.4 & 78.3 & {2} & 13.4  \\
			
			CenterCLIP~(spectral, $B_6-4, 49$)
			& 14.9 & 40.8 
			& 43.4 & 70.5 & 79.8 & {2} & {15.7} 
			& 42.1 & 70.5 & 80.6 & 2 & 11.7 \\
			
			CenterCLIP~(spectral, $B_6-3, 49$)
			& 14.2 & 43.6 
			& 43.7 & 71.3 & 80.2 & 2 & 16.2
			& 43.2 & 71.0 & 80.4 & 2 & 12.3  \\
			\midrule
			
			baseline (CLIP4clip~(meanP), ViT-B/16)
			& 25.7 & 59.6
			& 44.7 & 71.8 & 81.6 & 2 & 14.3
			& 46.5 & 73.4 & 82.3 & 2 & 11.0 \\			
			
			CenterCLIP~(k-medoids++, $B_6-4, 160$)
			& 17.6 & 86.5 
			& \cellcolor{Gray}47.5 & \cellcolor{Gray}74.4 & \cellcolor{Gray}82.5 & \cellcolor{Gray}2 &  \cellcolor{Gray}13.7 
			& \cellcolor{Gray}46.9 & \cellcolor{Gray}73.4 & \cellcolor{Gray}83.2 & \cellcolor{Gray}2 & \cellcolor{Gray}9.3 \\

			\bottomrule
		\end{tabular}
	}
	\caption{Training on \textsf{training-7K}}
	\end{subtable}%

	\begin{subtable}{1.0\linewidth}
	\centering
	\scalebox{0.83}{
		\begin{tabular}{lcc|ccccc|ccccc}
			\toprule 
			\multirow{2}{*}{\centering Method}
			& MeM.  & Speed
			& \multicolumn{5}{c}{Text $\rightarrow$ Video}
			& \multicolumn{5}{c}{Video $\rightarrow$ Text } \\

			~& GB & ms
			& R@1$\uparrow$   & R@5$\uparrow$ & R@10$\uparrow$ & MdR$\downarrow$ & MnR$\downarrow$ 
			& R@1$\uparrow$   & R@5$\uparrow$ & R@10$\uparrow$ & MdR$\downarrow$ & MnR$\downarrow$  \\
			\midrule 

			ActBERT~\cite{zhu2020actbert}
			& - & - & 8.6 & 23.4 & 33.1 & 36 & - 
			& - & - & - & - & - 			\\ 
			JSFusion~\cite{yu2018joint}	
			& - & - 
			& 10.2 & 31.2 & 43.2 & 13 & - 
			& - & - & - & - & - 		\\ 
			HowTo100M~\cite{miech2019howto100m}
			& - & - 
			& 14.9 	& 40.2 & 52.8 & 9 & - 
			& - 	& - & - & - & -\\ 
			CE~\cite{liu2019use}	
			& - & - 
			& 20.9 & 48.8 & 62.4 & 6 & - 
			& 20.6 & 50.3 & 64.0 & 5.3 & -\\ 
			MMT~\cite{gabeur2020multi}	
			& - & - 
			& 26.6 & 57.1 & 69.6 & 4 & 24.0 
			& 27.0 & 57.5 & 69.7 & 3.7 & 21.3\\
			T2VLAD~\cite{wang2021t2vlad}
			& - & - 
			& 29.5 & 59.0 & 70.1 & 4 & - 
			& 31.8 & 60.0 & 71.1 & 3 & - \\				
			AVLnet~\cite{rouditchenko2020avlnet}
			& - & - 
			& 27.1 & 55.6 & 66.6 & 4 & - 
			& 28.5 & 58.6 & 71.6 & 3 & - \\	
			TT-CE+~\cite{croitoru2021teachtext}
			& - & - 
			& 29.6 & 61.6 & 74.2 & 3 & - 
			& 32.1 & 62.7 & 75.0 & 3 & - \\			
	
			CLIP zero-shot
			 & - & - 
			& 31.2 & 53.7 & 64.2 & 4 & -
			& 27.2 & 51.7 & 62.6 & 5 & - \\
			
			CLIP4clip~(meanP)~\cite{2021clip4clip}
			& 20.8 & 24.4
			& 43.1 & 70.4 & 80.8 & {2} & 16.2 
			& 43.1 & 70.5 & 81.2 & {2} & 12.4 \\
			
			CLIP4clip~(seqTransf)
			& - & - 
			& {44.5} & {71.4} & {81.6} & {2} & 15.3
			& 42.7 & 70.9 & 80.6 & {2} & 11.6 \\
			\midrule 

			baseline (CLIP4clip~(meanP), ViT-B/32)
			& 20.8 & 24.4 
			& 43.0 & 70.7 & 80.6 & {2} & 16.2
			& 43.1 & 70.8 & 80.6 & {2} & 11.4 \\
			
			CenterCLIP~(k-medoids++, $B_6-4, 49$)
			& 15.0 & 22.9 
			& 43.6 & {71.4} & 81.2 & 2 & 15.3 
			& 42.9 & 70.4 & 80.8 & {2} & {10.8} \\
			
			CenterCLIP~(k-medoids++, $B_6-3, 49$)
			& \cellcolor{Gray}{14.2} & \cellcolor{Gray}{22.9}
			& 44.0 & 70.7 & 81.4 & {2} & 15.7
			& 42.9 & {71.4} & {81.7} & {2} & 11.1  \\

			CenterCLIP~(spectral, $B_6-4, 49$)
			& 14.9 & 40.8 
			& 43.6 & {71.7} & {80.6} & {2} & {15.4} 
			& 43.5 & 72.1	& 82.2 & 2 & 11.1 \\

			CenterCLIP~(spectral, $B_6-3, 49$)
			& 14.2 & 43.6 
			& 44.2 & 71.6 & 82.1 & 2 & 15.1
			& 42.8 & 71.7 & 82.2 & 2 & 10.9  \\
			\midrule

			baseline (CLIP4clip~(meanP), ViT-B/16)
			& 25.7 & 59.6
			& 45.6 & 71.2 & 80.9 & 2 & 15.2
			& 43.2 & 72.5 & 80.7 & 2 & 10.9 \\			
			
			CenterCLIP~(k-medoids++, $B_6-4, 160$)
			& 17.6 & 86.5 
			& \cellcolor{Gray}48.4 & \cellcolor{Gray}73.8 & \cellcolor{Gray}82.0 & \cellcolor{Gray}2 &  \cellcolor{Gray}13.8 
			& \cellcolor{Gray}47.7 & \cellcolor{Gray}75.0 & \cellcolor{Gray}83.3 & \cellcolor{Gray}2 & \cellcolor{Gray}10.2 \\

			\bottomrule
		\end{tabular}
	}
	\caption{Training on \textsf{training-9K}}
	\end{subtable}%

	\caption{ Results on MSR-VTT.
	MeM. is the average GPU memory cost when training on 2 and 8 Tesla V100 GPUs for ViT-B/32 and ViT-B/16, respectively.
	Speed is the inference time per video during evaluation on a Tesla V100 GPU. 
	}
	\label{tab:result_MSR-VTT}
\end{table*}


\noindent
\textbf{Learning strategies.}
We apply warm up and
cosine learning rate decay policy~\cite{2017_PriyaGoyal,2018_bags_of_tricks}.
If the initial learning rate is $lr$ and current epoch is $epoch$,
for the first $\textit{slow\_epoch}$ steps, the learning rate is 
$lr \times \frac{\textit{epoch}}{\textit{slow\_epoch}}$;
for the rest epochs, the learning rate is
$0.5 \times lr \times(1 + \cos ( \pi \times 
\frac{\textit{epoch} - \textit{slow\_epoch}}{\textit{max\_epoch} - \textit{slow\_epoch}} ) )$.
Generally, $lr$ is $\text{1e-5}$ for ActivityNet and  $\text{5e-6}$ for other datasets;
$\textit{max\_epoch}$ is 8 for ActivityNet and 5 for other datasets;
$\textit{slow\_epoch} = 0.1 \times \textit{max\_epoch}$.
AdamW~\cite{loshchilov2018decoupled} optimizer is adopted with
decoupled weight decay value 0.2.

\noindent
\textbf{Sequence length and batch size.}
For ActivityNet, the maximal text sequence length is 77 and
the frame length is 60.
For other datasets, the maximal text sequence length is 32 and
the frame length is 12.
The total batch size is always 128.
Experiments with ViT-B/32 for ActivityNet are done on 8 NVIDIA
Tesla V100 GPUs.
Experiments with ViT-B/32 for other datasets need at least 2 RTX 3090 GPUs.
All experiments are done with mixed precision~\cite{2018_AMP}.

\noindent
\textbf{Frame sampling.}
We adopt a sparse sampling strategy following TSN~\cite{wang2016temporal}.
During training, video frames are divided into $N_{in}$ 
segments, and we randomly sample one frame from each 
segment.
During the evaluation, $N_{in}$ frames are uniformly sampled from the video.
As the above said, $N_{in}= 60$ for ActivityNet
and $N_{in}= 12$ for the other datasets.
These $N_{in}$ frames are further divided into
$S$ segments during token clustering.

\noindent
\textbf{Evaluation metric.}
We use standard retrieval metrics: recall at rank $\mathbb{K}$
(R@$\mathbb{K}$, higher is better), median rank (MdR,
lower is better), and mean rank (MnR, lower is
better) to evaluate the performance.
R@$\mathbb{K}$ (Recall at $\mathbb{K}$) calculates the percentage of
test samples for which the correct result is found in
the top-$\mathbb{K}$ retrieved points to the query sample.
R@1, R@5, and R@10 are reported.
Median Rank calculates the
median of ground-truth results in the ranking.
Mean Rank calculates the average rank of
all correct results.

\begin{table*}[t]
	\centering
	\scalebox{0.82}{
		\begin{tabular}{lcc|ccccc|ccccc}
			\toprule 
			\multirow{2}{*}{\centering Method}
			& MeM.  & speed
			& \multicolumn{5}{c}{Text $\rightarrow$ Video}
			& \multicolumn{5}{c}{Video $\rightarrow$ Text } \\
			
			~& GB & ms
			& R@1$\uparrow$   & R@5$\uparrow$ & R@10$\uparrow$ & MdR$\downarrow$ & MnR$\downarrow$ 
			& R@1$\uparrow$   & R@5$\uparrow$ & R@10$\uparrow$ & MdR$\downarrow$ & MnR$\downarrow$  \\
			\midrule 
			
			JSFusion~\cite{yu2018joint}	
			& - & - 
			& 9.1 & 21.2 & 34.1 & 36.0 & - 
			& 12.3 & 28.6 & 38.9 & 20.0 & - 		\\ 

			CE~\cite{liu2019use}	
			& - & - 
			& 11.2 & 26.9 & 34.8 & 25.3 & 96.8 
			&  - & -  & -  & -  & -				\\

			MMT~\cite{gabeur2020multi}	
			& - & - 
			& 12.9 & 29.9 & 40.1 & 19.3 & 75.0 
			& 12.3 & 28.6 & 38.9 & 20.0 & 76.0 \\			
			
			Frozen in Time~\cite{bain2021frozen}
			& - & - 
			& 15.0 & 30.8 & 39.8 & 20.0 & - 
			& - & - & - & - & - \\		

			TT-CE+~\cite{croitoru2021teachtext}
			& - & - 
			& 17.2 & 36.5 & 46.3 & 13.7 & - 
			& 17.5 & 36.0 & 45.0 & 14.3 & - 	\\		

			CLIP zero-shot
			& - & - 
			& 15.1 & 28.3 & 35.8 & 31.0 & 132
			& 7.5 & 18.4 & 25.1 & 58.0 & 151 \\
			
			CLIP4clip~(meanP)~\cite{2021clip4clip}
			& 20.8 & 24.4
			& 20.7 & 38.9 & 47.2 & 13.0 & 65.3 
			& 20.6 & 39.4 & 47.5 & 13.0 & 56.7 \\
			
			CLIP4clip~(seqTransf)
			& - & - 
			& 22.6 & 41.0 & 49.1 & 11.0 & 61.0
			& 20.8 & 39.0 & 48.6 & 12.0 & 54.2 \\
			\midrule 

			baseline (CLIP4clip~(meanP), ViT-B/32)
			& 20.8 & 24.4
			& 20.1 & 40.2 & 48.4 & 12.0 & 57.1
			& 21.2 & 39.3 & 48.4 & 12.0 & 50.8 \\

			CenterCLIP~(k-medoids++, $B_6-6, 49$)
			& 16.4 & 23.9 
			& 21.9 & 41.1 & 50.7 & 10.0 & 55.6
			& 21.1 & 41.2 & 50.2 & 10.0 & 48.7  \\

			CenterCLIP~(k-medoids++, $B_6-4, 49$)
			& \cellcolor{Gray}15.0 & \cellcolor{Gray}22.9 
			& 21.7 & 39.8 & 49.8 & 11.0 & 54.8 
			& 21.4 & 40.3 & 50.8 & 10.0 & 48.4 	  \\
			
			CenterCLIP~(spectral, $B_6-6, 49$)
			& 16.4 & 40.8 
			& 21.6 & 40.9 & 49.3 & 11.0 & 57.2 
			& 20.6 & 39.5 & 48.8 & 12.0 & 51.4 \\
			
			CenterCLIP~(spectral, $B_6-4, 49$)
			& 15.0 & 43.6 
			& 21.4 & 39.7 & 49.4 & 11.0 & 55.9
			& 19.5 & 39.9 & 48.0 & 12.0 & 50.1  \\
			\midrule
			
			baseline (CLIP4clip~(meanP), ViT-B/16)
			& 25.7 & 59.6
			& 24.1 & 45.0 & 55.1 & 8 & 51.1
			& 22.5 & 42.9 & 53.5 & 9 & 45.1 \\			
			
			CenterCLIP~(k-medoids++, $B_6-4, 160$)
			& 17.6 & 86.5 
			& \cellcolor{Gray}24.2 & \cellcolor{Gray}46.2 & \cellcolor{Gray}55.9 & \cellcolor{Gray}8 &  \cellcolor{Gray}47.3 
			& \cellcolor{Gray}24.5 & \cellcolor{Gray}46.4 & \cellcolor{Gray}55.8 & \cellcolor{Gray}7 & \cellcolor{Gray}41.3 \\
			\bottomrule
		\end{tabular}
	}
	\caption{ Results on LSMDC. 
		MeM. is the average GPU memory cost when training on 2 and 8 Tesla V100 GPUs for ViT-B/32 and ViT-B/16, respectively.
		Speed is the inference time per video during evaluation on a Tesla V100 GPU. 
	}
	\label{tab:result_LSMDC}
\end{table*}

\noindent
\textbf{Setting of CenterCLIP.}
We use {\color{blue} {$\bm{(B_a-S, K)}$}} to represent the setting.
It means we perform token clustering right after the $a$-th transformer
block, the number of temporal segments is $S$,
and the number of clusters/centers
are constant $K$.
Generally, we construct KNN graph with Gaussian similarity function
between two points when applying spectral clustering: $\exp(-\lVert x_i - x_j \rVert^2 / (2\sigma^2))$.
The neighbours of one vertex is $5 \times$ \{the number of frames in a segment\} for ViT-B/32,
and plus an additional 5 for ViT-B/16.
The variance of the Gaussian function $\sigma$ is simply set to 2.0.
No normalization is applied for token embeddings before performing clustering.
Baselines in the
experiments use the same setting
as CenterCLIP.

\subsection{Results on Common Benchmarks}
As shown in Table~\ref{tab:result_MSVD}, Table~\ref{tab:result_ACT}, 
Table~\ref{tab:result_MSR-VTT}, and Table~\ref{tab:result_LSMDC}.
We achieve SOTA performance on all four datasets.
Moreover, we also achieve decent memory usage reduction in all
cases and obvious speedup of evaluation in some cases.
Specifically, for CenterCLIP~(ViT-B/32),
we achieve a 32\% reduction in memory cost and accelerate the
model by 6\% of the original speed for MSR-VTT, MSVD, and LSMDC in the best situation.
For ActivityNet, the reduction in memory cost is 35\% and the speedup of evaluation
speed is 14\% in the best case.
These numbers verify the efficiency of our method.
For CenterCLIP~(ViT-B/16),
as the patch size decreases, the number of tokens increases
(4 $\times$ as the number of tokens of ViT-B/32).
In this work, 
the clustering complexity is at least linearly related to the number
of data points.
Therefore, CenterCLIP does not gain speedup for ViT-B/16.
However, CenterCLIP also achieves a 32\% reduction in memory cost.
In future work, we will introduce faster clustering algorithms
to speed up the whole model.

Compared to the baseline, CenterCLIP
achieves significant improvement on recall.
When using ViT-B/32,
for MSVD, the maximal gain of text$\rightarrow$video R@1 is 1.7\%;
for MSR-VTT (\textsf{training-9K}), the number is 1.2\%;
for LSMDC, it is 1.8\%;
for ActivityNet, it achieves  2.1\% improvement of text$\rightarrow$video R@1.
When using ViT-B/16,
for MSVD, the numbers are 1.0\%, 2.8\%, and 0.1\% for text$\rightarrow$video R@1,
5.7\%, 5.2\%, and 2.0\% for video$\rightarrow$text R@1.
CenterCLIP gains more improvement of video$\rightarrow$text retrieval performance in this case.
All these results demonstrate the effectiveness of our clustering strategy.
It aligns segment semantics of text and video.

It is worth noting that spectral clustering and k-medoids++ achieve similar performance
in most cases.
This is somehow counter-intuitive as spectral clustering should be more suitable
for clustering high-dimensional points.
This is possible because the data shape of clusters of token embeddings in high dimensional space is
nearly spherical.
Spectral clustering does achieve better performance in terms of some metrics,
\eg, better R@5 and R@10 on MSR-VTT and ActivityNet,
and produces the best video$\rightarrow$text results on MSVD.

\subsection{Diagnostic Experiments}
In this section, we will analyze CenterCLIP thoroughly.
All diagnostic experiments are taken with CenterCLIP~(ViT-B/32).

\begin{table}[htbp]
	\centering
	\resizebox{0.47\textwidth}{!}{
		\begin{tabular}{lcc|ccccc}
			\toprule 

			Method & Mem. & T$\leftrightarrow $V
			& R@1$\uparrow$   & R@5$\uparrow$ & R@10$\uparrow$ & MdR$\downarrow$ & MnR$\downarrow$
			\\
			\midrule
			
			\multicolumn{8}{c}{MSR-VTT (train on \textsf{training-7K})}
			\\
			\midrule

			\multirow{2}{*}{\centering pooling ($B_6-{6}, 49$)}
			& \multirow{2}{*}{\centering 16.39} & T$\rightarrow$V
			& 41.9 & 66.6 & 76.7 & 2 & 18.7 \\
			
			~
			&~ & V$\rightarrow$T
			& 40.2 & 65.6 & 75.8 & 2 & 14.4 \\
			
			\multirow{2}{*}{\centering pooling ($B_6-{4}, 49$)}
			& \cellcolor{Gray1} & T$\rightarrow$V
			& 40.6 & 65.6 & 75.8 & 2 & 17.5 \\

			~
			&\cellcolor{Gray1}\multirow{-2}{*}{\centering  14.95} & V$\rightarrow$T
			& 40.6 & 67.3 & 77.3 & 2 & 14.6 \\

			\multirow{2}{*}{\centering sparse sampling ($B_6-{6}, 49$)}
			& \multirow{2}{*}{\centering 16.39} & T$\rightarrow$V
			& 42.6 & 68.4 & 78.4 & 2 & 17.6 \\
			
			~
			&~ & V$\rightarrow$T
			& 41.6 & 68.3 & 77.5 & 2 & 12.8 \\
			
			\multirow{2}{*}{\centering sparse sampling ($B_6-{4}, 49$)}
			& \cellcolor{Gray1} & T$\rightarrow$V
			& 42.3 & 69.1 & 78.6 & 2 & 17.6 \\

			~
			&\cellcolor{Gray1} \multirow{-2}{*}{\centering  14.95} & V$\rightarrow$T
			& 40.3 & 66.7 & 77.0 & 2 & 13.5 \\

			\multirow{2}{*}{\centering token shift~\cite{zhang2021token}}
			& \multirow{2}{*}{\centering 20.77} & T$\rightarrow$V
			& 42.5 & 68.5 & 79.6 & 2 & \cellcolor{Gray} 16.4 \\
			
			~
			&~ & V$\rightarrow$T
			& \cellcolor{Gray1}43.3 & 70.1 & \cellcolor{Gray1}80.8 & 2 & 12.2 \\

			\multirow{2}{*}{\centering temporal shift~\cite{wang2016temporal}}
			& \multirow{2}{*}{\centering 20.77} & T$\rightarrow$V
			& 34.2 & 61.8 & 73.7 & 3 & 21.9 \\
			
			~
			&~ & V$\rightarrow$T
			& 31.5 & 61.8 & 72.2 & 3 & 18.2 \\	
			
			\multirow{2}{*}{\centering CenterCLIP ($B_6-{6}, 49$)}
			& \multirow{2}{*}{\centering 16.39} & T$\rightarrow$V
			& 43.3 & 69.9 & 78.6 & 2 & 17.7 \\
			
			~
			&~ & V$\rightarrow$T
			& 41.8 & 68.9 & 77.3 & 2 & 12.7 \\
			
			\multirow{2}{*}{\centering CenterCLIP ($B_6-{4}, 49$)}
			&\cellcolor{Gray1}  & T$\rightarrow$V
			& \cellcolor{Gray}43.7 & \cellcolor{Gray}71.3 & \cellcolor{Gray}80.8 & 2 & 16.9 \\
			
			~
			& \cellcolor{Gray1} \multirow{-2}{*}{\centering  14.95} & V$\rightarrow$T
			& 41.8 & \cellcolor{Gray1}70.2 & 79.8 & 2 & \cellcolor{Gray1}11.8 \\

			\midrule
			\multicolumn{8}{c}{LSMDC}
			\\
			\midrule
			\multirow{2}{*}{\centering sparse sampling ($B_6-{6}, 49$)}
			& \multirow{2}{*}{\centering 16.39} & T$\rightarrow$V
			& 20.6 & 38.7 & 48.6 & 12.0 & 59.8 \\
			
			~
			&~ & V$\rightarrow$T
			& 20.4 & 37.9 & 45.6 & 13.5 & 54.2 \\			

			\multirow{2}{*}{\centering token shift~\cite{zhang2021token}}
			& \multirow{2}{*}{\centering 20.77} & T$\rightarrow$V
			& 21.4 & \cellcolor{Gray}42.3 & \cellcolor{Gray}50.2 & \cellcolor{Gray}10.0 & 55.6 \\
			
			~
			&~ & V$\rightarrow$T
			& \cellcolor{Gray1}21.7 & \cellcolor{Gray1}41.6 &  50.1 & 10.0 & 49.6 \\	

			\multirow{2}{*}{\centering CenterCLIP ($B_6-{4}, 49$)}
			& \cellcolor{Gray1}  & T$\rightarrow$V
			& \cellcolor{Gray}21.7 & 39.8 & 49.8 & 11.0 & \cellcolor{Gray}54.8 \\
			 
			~
			&\cellcolor{Gray1} \multirow{-2}{*}{\centering  14.95} & V$\rightarrow$T
			& 21.4 & 40.3 & \cellcolor{Gray1}50.8 & 10.0 & \cellcolor{Gray1}48.4 \\		

			\midrule
			\multicolumn{8}{c}{ActivityNet} \\
			\midrule

			\multirow{2}{*}{\centering token shift~\cite{zhang2021token}}
			& \multirow{2}{*}{\centering 24.98} & T$\rightarrow$V
			& 42.0 & 73.6 & 84.8 & 2.0 & 7.3 \\

			~
			& ~ & V$\rightarrow$T
			& 42.5 & 74.3 &  85.2 & 2.0 & 7.0 \\	

			\multirow{2}{*}{\centering CenterCLIP ($B_6-{15}, 49$)}
			& \cellcolor{Gray1}  & T$\rightarrow$V
			& \cellcolor{Gray}43.9 & \cellcolor{Gray}75.3 & \cellcolor{Gray}85.2 & 2.0 & \cellcolor{Gray}7.0 \\

			~
			&\cellcolor{Gray1}\multirow{-2}{*}{\centering  16.75} & V$\rightarrow$T
			& \cellcolor{Gray1}44.2 & \cellcolor{Gray1}75.0 & \cellcolor{Gray1}86.1 & 2.0 & \cellcolor{Gray1}6.8 \\				
			\bottomrule
		\end{tabular}
	}
	\caption{Comparison with strong token selection baselines. 
	}
	\label{tab:more_baseline}
\end{table}

\begin{figure}[htbp]
	\centering
	\begin{subfigure}{.235\textwidth}
		\centering
		\includegraphics[width=1.0\textwidth]{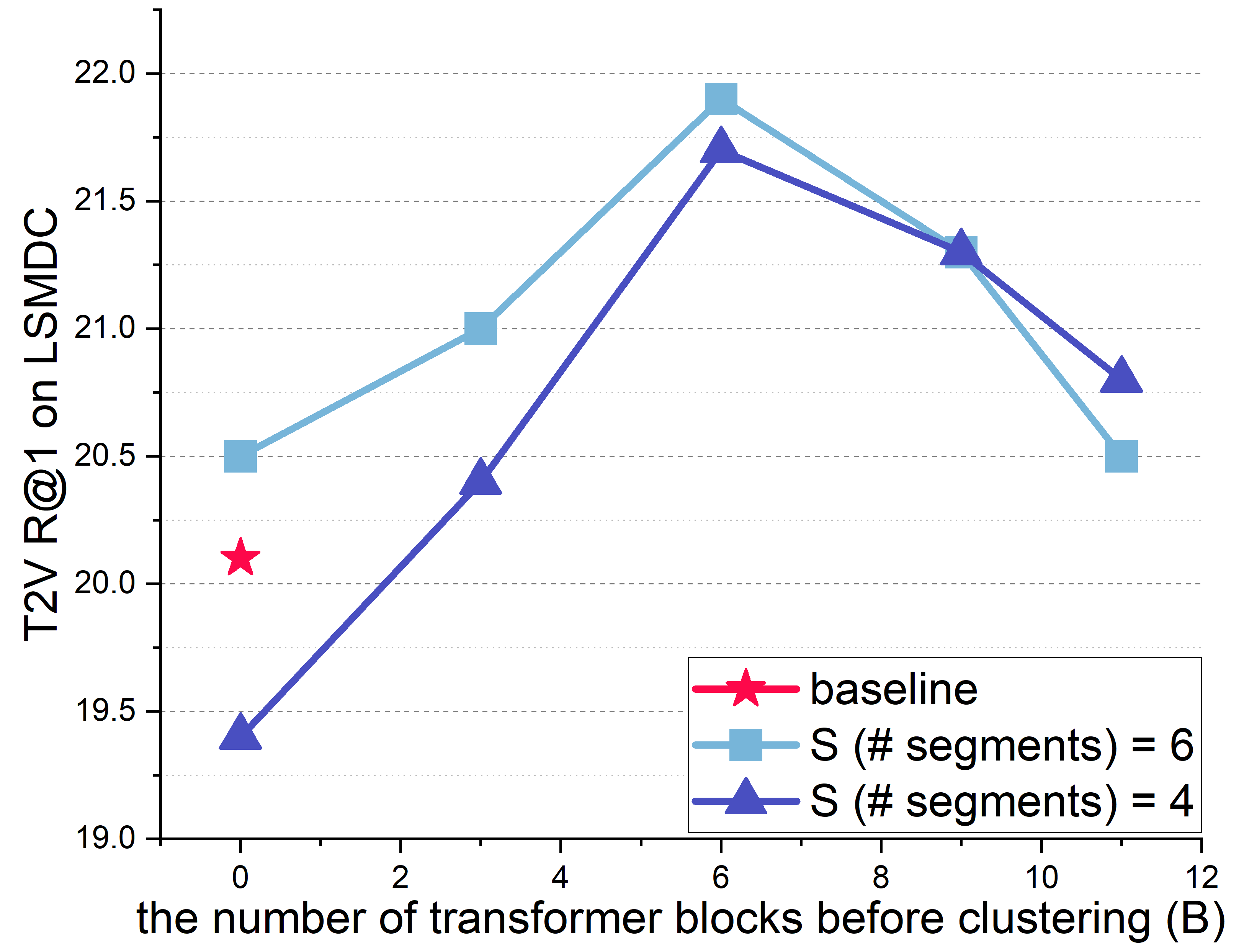} 
		\caption{T2V R@1 on LSMDC}
		\label{fig:sub1}
	\end{subfigure}%
	\begin{subfigure}{.235\textwidth}
		\centering
		\includegraphics[width=1.0\textwidth]{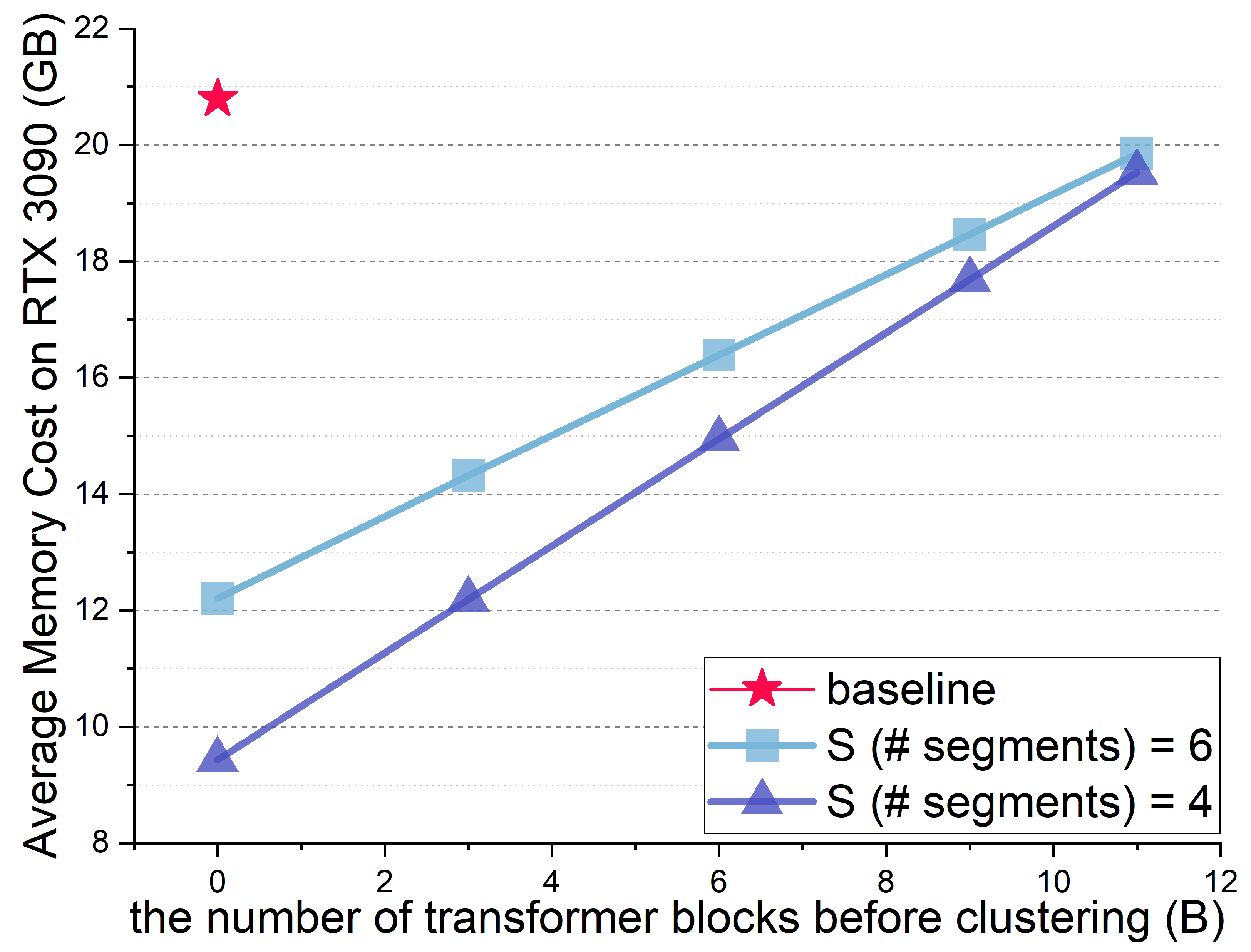} 
		\caption{Memory cost on RTX 3090}
		\label{fig:sub2}
	\end{subfigure}

	\caption{Influence of places of token clustering.}
	\label{fig_place}
\end{figure}

\subsubsection{More baselines}
We provide four more strong baselines:
~\textit{1)} pooling of nearby tokens in a temporal segment,
after pooling, we get $K$ average tokens for one segment;
~\textit{2)} sparse sampling of tokens in a temporal segment,
namely, randomly sample
$K$ tokens from a temporal segment during training and
uniformly sample $K$ tokens during validation;
~\textit{3)} temporal shift described in TSM~\cite{wang2016temporal},
here we apply temporal shift to the
tokens except [\texttt{CLASS}] embedding;
~\textit{4)} token shift described in~\cite{zhang2021token},
the method only shift the [\texttt{CLASS}] embedding.
The shift is performed twice in each transformer block,
right before MHSA and FFN.
Results are shown in Table~\ref{tab:more_baseline}.
Shifting all image patches does not work here.
sparse sampling produces a little better results than baseline on
MSR-VTT and LSMDC.
However, CenterCLIP is much better than sparse sampling,
this demonstrates the necessity of selecting representative tokens.
Tokenshift achieves pretty good performance on short videos,
nevertheless, it does not reduce any computational costs.

\subsubsection{The place of performing token clustering}
The influence of places of token clustering (k-medoids++)
is shown in Figure~\ref{fig_place}.
The smaller the $B$, the lower the memory cost.
The performance of the whole model will also decrease along
with the decreasing or increasing of $B$.
A good trade-off between memory cost and performance achieves at $B=6$,
this is also our default setting.

We can also take multiple times of clustering.
For instance, firstly perform clustering with $(S=6, B=4)$
and then with $(S=3, B=8)$.
The results are shown in Table~\ref{tab:twice_cluster}.
Such progressive clustering strategy achieves pretty good R@5,
R@10, MdR, and memory cost reduction.
However, performing multiple times will increase the time complexity
and this is not suitable for large amounts of tokens.
Thus we generally perform clustering once in this work.

\begin{table}[tbp]
	\centering
	\resizebox{0.475\textwidth}{!}{
		\begin{tabular}{lcc|ccccc}
			\toprule 

			Method & Mem. & T$\leftrightarrow $V
			& R@1$\uparrow$   & R@5$\uparrow$ & R@10$\uparrow$ & MdR$\downarrow$ & MnR$\downarrow$
			\\
			\midrule

			CenterCLIP
			& \cellcolor{Gray1}~ & T$\rightarrow$V
			& 21.0 & \cellcolor{Gray}42.7 & \cellcolor{Gray}51.9 & \cellcolor{Gray}9.0 & 57.0 \\
			
			($B_4-{6}, 49$), ($B_8-{3}, 49$)
			& \cellcolor{Gray1}  \multirow{-2}{*}{\centering 13.52} & V$\rightarrow$T
			& 20.9 & \cellcolor{Gray1}41.8 & \cellcolor{Gray1} 51.5 & \cellcolor{Gray1}9.0 & 50.7 \\

			\multirow{2}{*}{\centering CenterCLIP ($B_6-{4}, 49$)}
			& ~  & T$\rightarrow$V
			& 21.7 & 39.8 & 49.8 & 11.0 &  \cellcolor{Gray}54.8 \\

			~
			& \multirow{-2}{*}{\centering  16.39} & V$\rightarrow$T
			& \cellcolor{Gray1}21.4 & 40.3 &  50.8 & 10.0 & \cellcolor{Gray1}48.4 \\		

			\multirow{2}{*}{\centering CenterCLIP ($B_6-{6}, 49$)}
			& ~  & T$\rightarrow$V
			& \cellcolor{Gray}21.9 & 41.1 & 50.7 & 10.0 & 55.6 \\
			 
			~
			& \multirow{-2}{*}{\centering  14.95} & V$\rightarrow$T
			& 21.1 & 41.2 & 50.2 & 10.0 & 48.7 \\
			
			\bottomrule
		\end{tabular}
	}
	\caption{Performing clustering twice on LSMDC.}
	\label{tab:twice_cluster}
\end{table}

\begin{figure}[tbp]
	\centering
	\begin{subfigure}{.235\textwidth}
		\centering
		\includegraphics[width=1.0\textwidth]{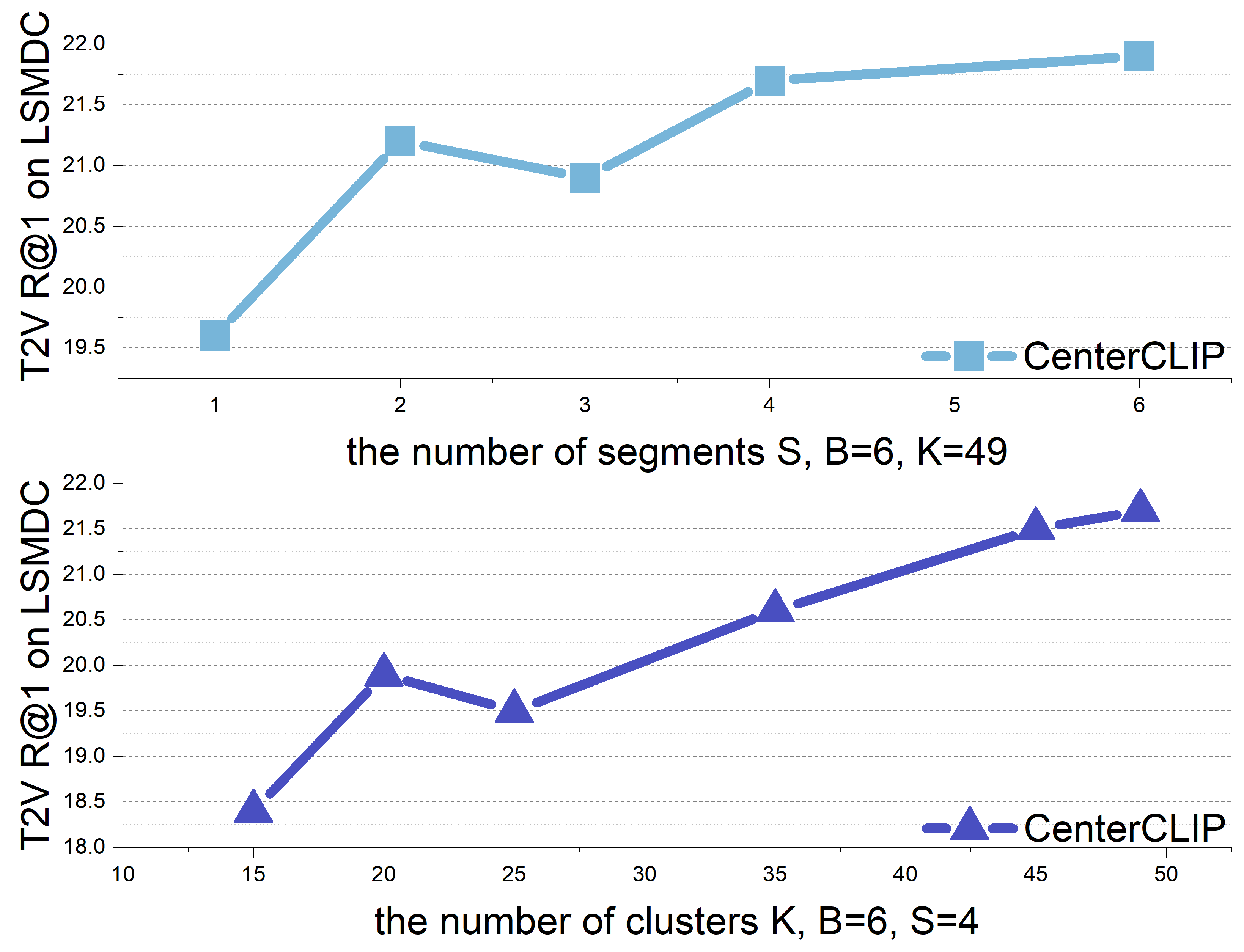} 
		\caption{T2V R@1 on LSMDC}
		\label{fig_ks_1}
	\end{subfigure}%
	\begin{subfigure}{.235\textwidth}
		\centering
		\includegraphics[width=1.0\textwidth]{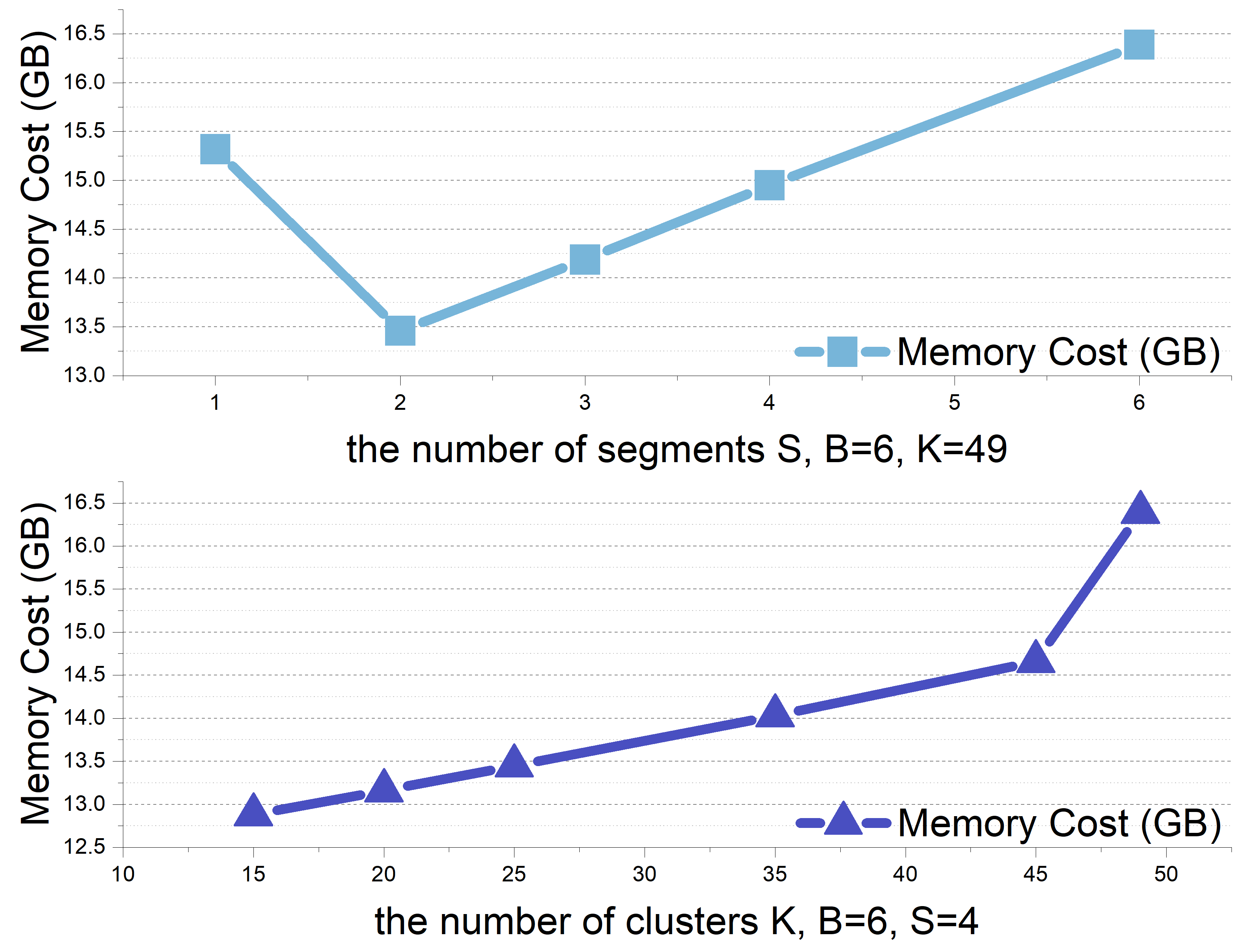} 
		\caption{Memory cost on RTX 3090}
		\label{fig_ks_2}
	\end{subfigure}
	\begin{subfigure}{.235\textwidth}
	\centering
	\includegraphics[width=1.0\textwidth]{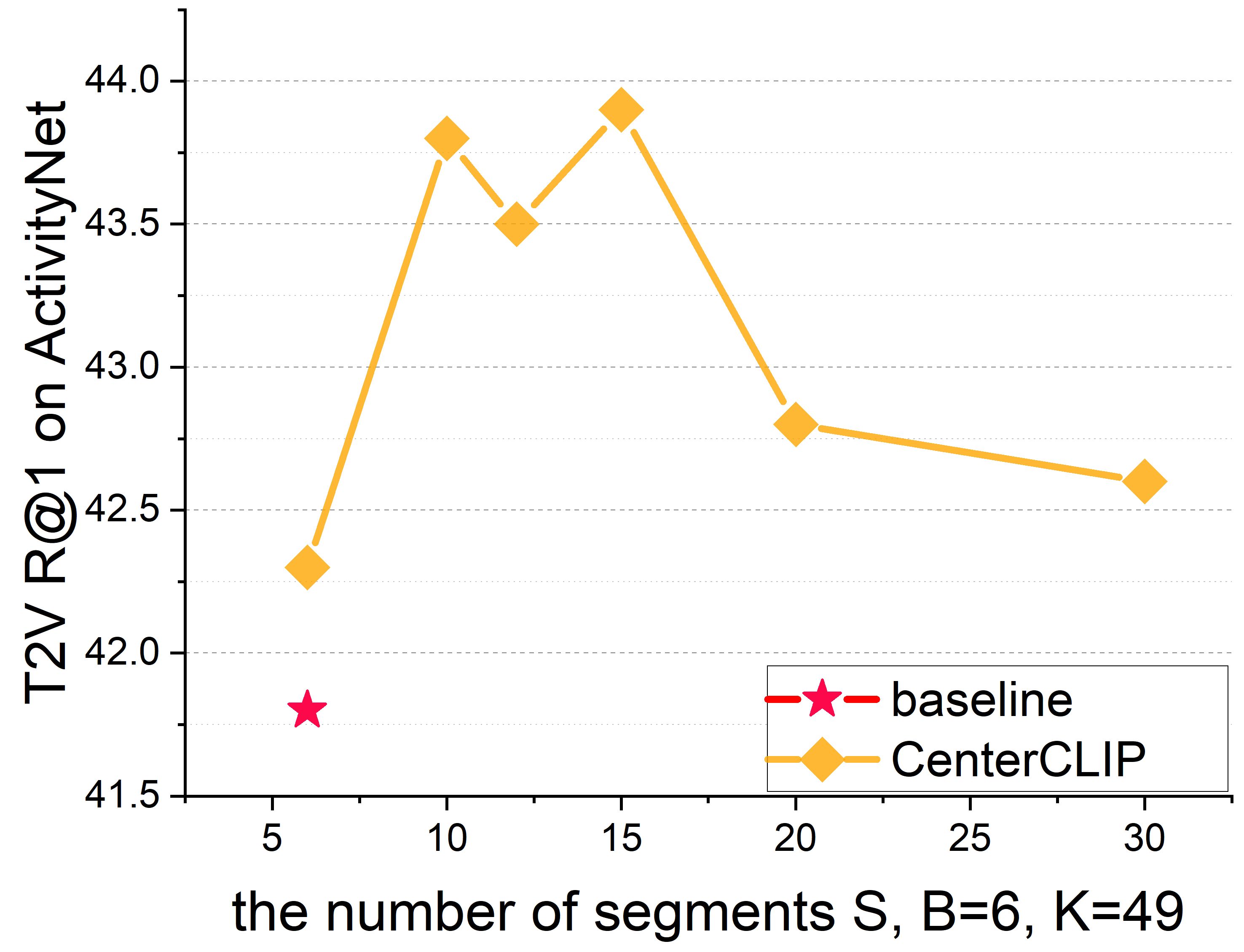} 
	\caption{T2V R@1 on ActivityNet}
	\label{fig_ks_3}
	\end{subfigure}
	\begin{subfigure}{.235\textwidth}
	\centering
	\includegraphics[width=1.0\textwidth]{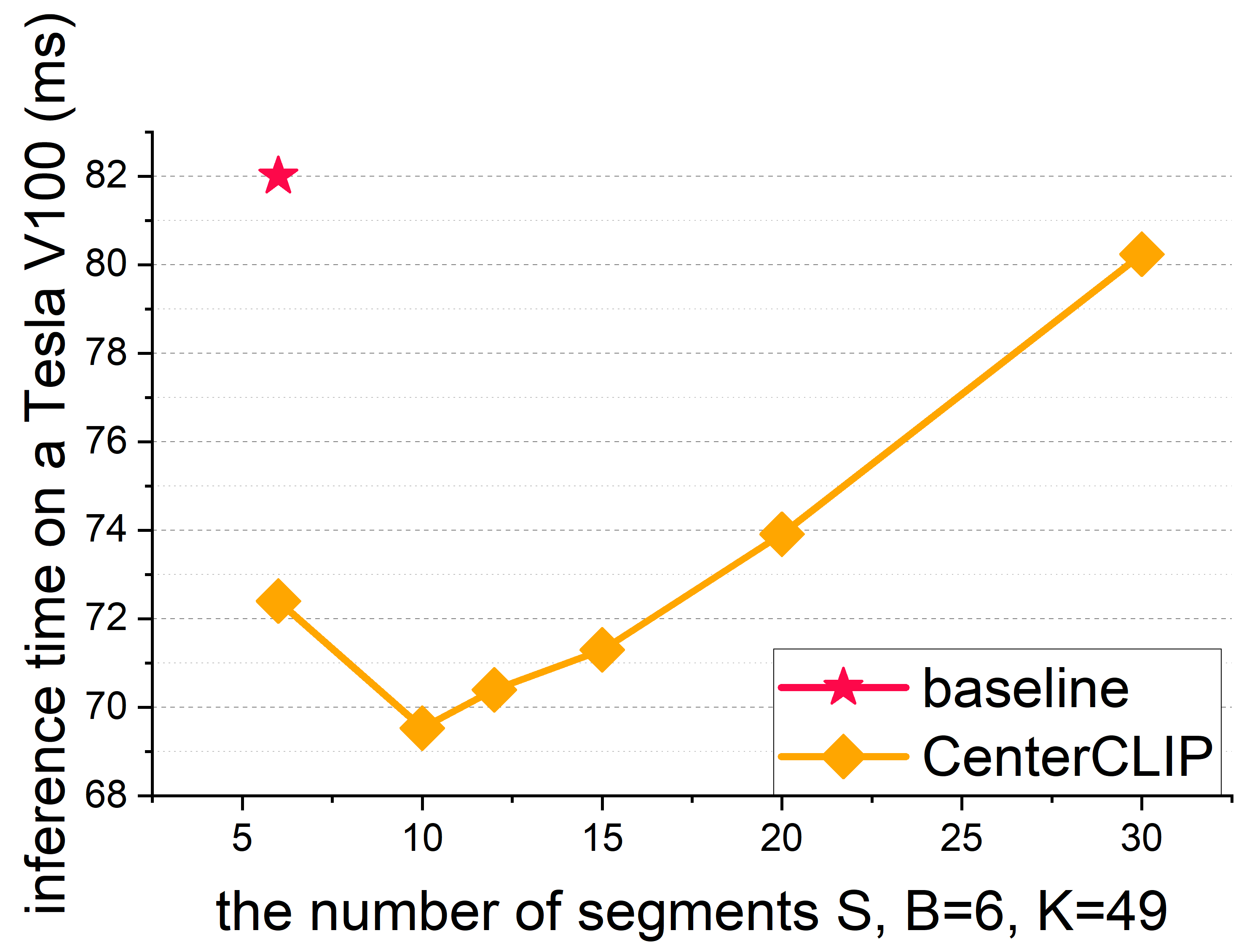} 
	\caption{Inference time}
	\label{fig_ks_4}
	\end{subfigure}	
	\caption{Influence of cluster number $K$ and segment $S$.}
	\label{fig_ks}
\end{figure}

\subsubsection{The number of cluster $K$ and segment $S$}
We perform experiments on LSMDC and ActivityNet.
The results including R@1, memory cost, and inference time
are shown in Figure~\ref{fig_ks}.
Along with the increase of $K$, the performance increases, and computation costs also increase.
At the same time, a small segment number $S$ does not always achieve
better performance, \eg, $S=1$ on LSMDC and $S=6$ on ActivityNet.
A small segment number means more tokens are dropped.
This will cause the loss of more information.
When $S=1$ on LSMDC and $S=6$ on ActivityNet, the number of tokens
in a segment is large, \ie, $12 \times 49$ and $10 \times 49$,
this leads to more computational costs of clustering as shown in Figure~\ref{fig_ks_3}
and Figure~\ref{fig_ks_4}.
Thus a moderate segment number $S$ is usually adopted.
 
\subsubsection{The number of input frames $N_{in}$}
We change the number of input video frames and take experiments
with CenterCLIP ($B_6-{15}, 49$) on ActivityNet.
The results are shown in Table~\ref{tab:in_frames}.
The large the number of input frames $N_{in}$,
the more computation costs,
and a small number of frames will lead to worse performance.
Similar ablations about the input of frames on short video
datasets like MSR-VTT can be found in CLIP4clip~\cite{2021clip4clip}.
When the number of segments $S$ is fixed, a large number
of input frames $N_{in}$ will also increase computation 
costs of the clustering process as the number of tokens in one temporal segment increases.

\begin{table}[tbp]
	\centering
	\resizebox{0.475\textwidth}{!}{
		\begin{tabular}{ccc|ccccc}
			\toprule 

			CenterCLIP ($B_6-{15}, 49$) & Mem. & T$\leftrightarrow $V
			& R@1$\uparrow$   & R@5$\uparrow$ & R@10$\uparrow$ & MdR$\downarrow$ & MnR$\downarrow$
			\\
			\midrule

			\multirow{2}{*}{\centering $N_{in}=75$}
			& ~  & T$\rightarrow$V
			& 43.8 & 74.8 & \cellcolor{Gray}85.8 & 2.0 & \cellcolor{Gray}6.7 \\
			 
			~
			& \multirow{-2}{*}{\centering  19.48} & V$\rightarrow$T
			& \cellcolor{Gray1}44.2 & \cellcolor{Gray1}75.5 & \cellcolor{Gray1}86.8 & 2.0 & \cellcolor{Gray1}6.5 \\

			\multirow{2}{*}{\centering $N_{in}=60$}
			& ~ & T$\rightarrow$V
			& \cellcolor{Gray}43.9 & \cellcolor{Gray}75.3 & 85.2 & 2.0 & 7.0 \\

			~
			&\multirow{-2}{*}{\centering  16.75} & V$\rightarrow$T
			& \cellcolor{Gray1}44.2 & 75.0 & 86.1 & 2.0 & 6.8 \\	

			\multirow{2}{*}{\centering $N_{in}=45$}
			& ~  & T$\rightarrow$V
			& 42.6 & 74.5 & 84.8 & 2.0 & 7.0 \\
			 
			~
			& \multirow{-2}{*}{\centering  14.03} & V$\rightarrow$T
			& 43.7 & 74.5 & 85.4 & 2.0 & 6.7 \\		

			\multirow{2}{*}{\centering $N_{in}=30$}
			& ~  & T$\rightarrow$V
			& 43.2 & 74.5 & 85.1 & 2.0 & 6.9 \\
			 
			~
			& \multirow{-2}{*}{\centering  11.29} & V$\rightarrow$T
			& 43.4 & 74.7 & 85.5 & 2.0 & 6.7 \\
			
			\bottomrule
		\end{tabular}
	}
	\caption{Influence of input frames $N_{in}$ on ActivityNet.}
	\label{tab:in_frames}
\end{table}

\begin{table}[tbp]
	\centering
	\resizebox{0.475\textwidth}{!}{
		\begin{tabular}{lcc|ccccc}
			\toprule 

			Method & Mem. & T$\leftrightarrow $V
			& R@1$\uparrow$   & R@5$\uparrow$ & R@10$\uparrow$ & MdR$\downarrow$ & MnR$\downarrow$
			\\
			\midrule

			CenterCLIP ($B_6-{4}, 49$)
			& ~ & T$\rightarrow$V
			& 21.6 & 40.5 & \cellcolor{Gray}50.9 & 10.0 & 55.8 \\
			
			\textbf{w.} $\ell$2 norm
			& \multirow{-2}{*}{\centering 16.39} & V$\rightarrow$T
			& \cellcolor{Gray1}22.1 & 40.8 & 50.5 & 10.0 & \cellcolor{Gray1}48.4 \\

			CenterCLIP ($B_6-{6}, 49$)
			& ~ & T$\rightarrow$V
			& 21.5 & 40.8 & 50.0 & 10.5 & 56.7 \\
			
			\textbf{w.} $\ell$2 norm
			& \multirow{-2}{*}{\centering 14.95} & V$\rightarrow$T
			& 20.2 & 39.8 & 48.6 & 12.0 & 50.5 \\

		    CenterCLIP ($B_6-{4}, 49$)
			& ~  & T$\rightarrow$V
			& 21.7 & 39.8 & 49.8 & 11.0 & \cellcolor{Gray}54.8 \\

			\textbf{wo.} $\ell$2 norm
			& \multirow{-2}{*}{\centering  16.39} & V$\rightarrow$T
			& 21.4 & 40.3 &  \cellcolor{Gray1}50.8 & 10.0 & 48.4 \\		

			CenterCLIP ($B_6-{6}, 49$)
			& ~  & T$\rightarrow$V
			& \cellcolor{Gray}21.9 & \cellcolor{Gray}41.1 & 50.7 & \cellcolor{Gray}10.0 & 55.6 \\
			 
			\textbf{wo.} $\ell$2 norm
			& \multirow{-2}{*}{\centering  14.95} & V$\rightarrow$T
			& 21.1 & \cellcolor{Gray1}41.2 & 50.2 & \cellcolor{Gray1}10.0 & 48.7 \\
			
			\bottomrule
		\end{tabular}
	}
	\caption{Performing k-medoids++ clustering with and without $\ell$2 normalization on LSMDC.}
	\label{tab:norm}
\end{table}

\begin{table}[tbp]
	\centering
	\resizebox{0.475\textwidth}{!}{
		\begin{tabular}{ccc|ccccc}
			\toprule 

			$lr$ & epochs & T$\leftrightarrow $V
			& R@1$\uparrow$   & R@5$\uparrow$ & R@10$\uparrow$ & MdR$\downarrow$ & MnR$\downarrow$
			\\
			\midrule

			\multirow{2}{*}{\centering $\text{5e-6}$}
			& ~  & T$\rightarrow$V
			& 40.6 & 72.3 & 83.7 & 2.0 & 7.7 \\
			 
			~
			& \multirow{-2}{*}{\centering  5} & V$\rightarrow$T
			& 41.8 & 73.1 & 84.5 & 2.0 & 7.0 \\

			\multirow{2}{*}{\centering $\text{1e-5}$}
			& ~ & T$\rightarrow$V
			& 41.0 & 73.4 & 84.2 & 2.0 & 7.3 \\

			~
			&\multirow{-2}{*}{\centering  5} & V$\rightarrow$T
			& 42.8 & 73.8 & 85.2 & 2.0 & 6.9 \\	

			\multirow{2}{*}{\centering $\text{1e-5}$}
			& ~  & T$\rightarrow$V
			& \cellcolor{Gray}41.8 & \cellcolor{Gray}73.9 & \cellcolor{Gray}84.7 & 2.0 & 7.3 \\
			~
			& \multirow{-2}{*}{\centering  8} & V$\rightarrow$T
			& \cellcolor{Gray1}42.8 & \cellcolor{Gray1}73.8 & \cellcolor{Gray1}85.3 & 2.0 & 6.9 \\
			
			\bottomrule
		\end{tabular}
	}
	\caption{Different learning rates and training epochs for CLIP4clip baselines on ActivityNet.}
	\label{tab:lre}
\end{table}

\subsubsection{Normalization of token embeddings}
People may be curious about the influence of embedding
normalization when performing token clustering.
We showed results with and without $\ell$2 normalization in Table~\ref{tab:norm}.
The difference is trivial.
Indeed, different choices of normalization are equal
to different choices of distance metrics in clustering.
Besides distance metrics, some other 
factors can also influence the results of 
clustering, \ie, how to construct graphs in spectral clustering.
However, this is not the focus of our work, and we do not make further explorations in this direction.

\begin{figure*}[!t]
	\includegraphics[width=0.985\textwidth]{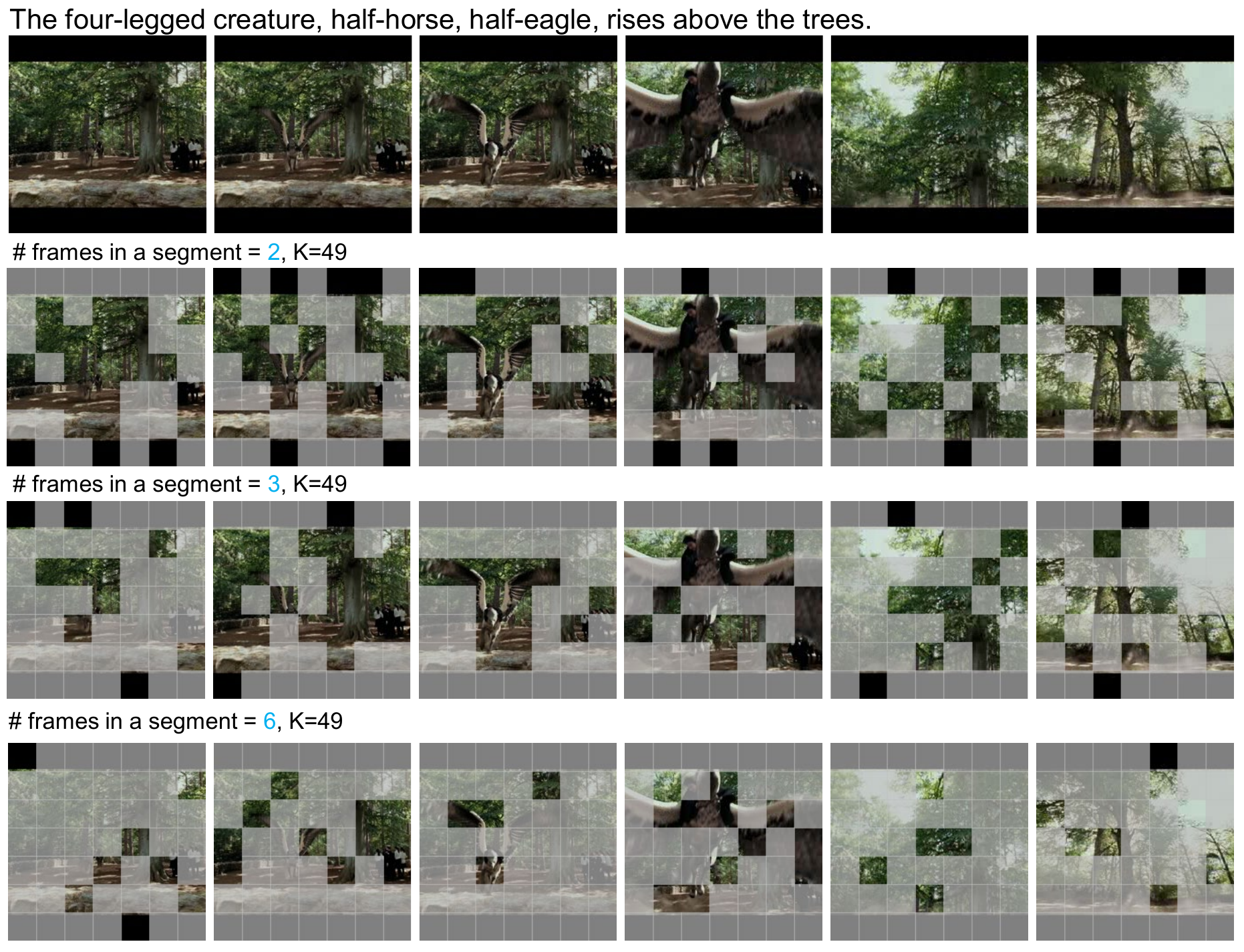}
	\caption{Visualization of centers after token clustering with different number of frames in a temporal segment.}
	\label{fig:vis}
\end{figure*}

\subsubsection{Learning rate and training epochs}
The original CLIP4clip uses a learning rate of 1e-7.
When setting $lr$ = $\text{1e-7}$ on MSR-VTT (\textsf{training-7K}), we get 39.7
T$\rightarrow$V R@1 with mixed precision~\cite{2018_AMP}
and 41.7 T$\rightarrow$V R@1 without mixed precision on the split `\textsf{test 1k-A}'.
The corresponding result of CLIP4clip baseline is 42.1 T$\rightarrow$V R@1.
When increasing the learning rate to $\text{5e-6}$, we get 
42.4 T$\rightarrow$V R@1 with mixed precision on `\textsf{test 1k-A}'.
As the mixed precision training saves a lot of GPU memory and
accelerates the training procedure,
we are stuck with mixed precision and 
use $lr=$ $\text{5e-6}$ for short video
datasets.
For ActivityNet, we found a large learning rate with more
training epochs brings a better result.
This is shown in Table~\ref{tab:lre}.
It is possibly because of the large number of different video frames in long videos in ActivityNet and the sparse sampling strategy we used during training.
The model needs more training epochs to 
learn good representations of videos.

\subsubsection{Visualization of image patches of center tokens after clustering}
We further display visualization results of center tokens
after multi-segment clustering with different numbers of video frames
within a temporal segment.
The results are shown in Figure~\ref{fig:vis}.
It is clear that the clustering algorithm reserves the most representative
tokens, for example, in the second and third row of Figure~\ref{fig:vis},
tokens of the foreground animal are selected and only part of the tokens of the similar background remains.
This verifies our beginning motivation that using a few typical tokens is already enough
for learning discriminative features for video representation.

\section{Conclusion}
In this work, we propose a multi-segment clustering algorithm to 
reduce the number of redundant tokens of continuous video frames,
and achieve segment-level alignment of video and text representations
for text-video retrieval task.
Our method, named CenterCLIP as we only reserve center tokens of token clusters
and drop non-center tokens,
is based on the knowledge of large-scale image-text pairs
pre-trained model -- CLIP.
We take extensive experiments on four common text-video multi-modal
datasets: MSR-VTT, MSVD, LSMDC, and ActivityNet.
CenterCLIP achieves state-of-the-art performance on all these four datasets
and surpass the old SOTA by a large margin.
At the same time, CenterCLIP realizes a decent reduction in memory costs and
speedup of inference time.

\section*{Acknowledgments}
This work is supported by National Key R\&D Program of China under Grant No. 2020AAA0108800.
Thanks Naiyuan Liu for his helpful discussions.

%

\newpage
\bibliographystyle{ACM-Reference-Format}
\bibliography{sample-base}


\end{document}